\documentclass{article}

     \PassOptionsToPackage{numbers, compress}{natbib}

\usepackage[preprint]{neurips_2020}

\usepackage[utf8]{inputenc} 
\usepackage[T1]{fontenc}    
\usepackage{hyperref}       
\usepackage{url}            
\usepackage{booktabs}       
\usepackage{amsfonts}       
\usepackage{amssymb} 
\usepackage{nicefrac}       
\usepackage{microtype}      
\usepackage[table,xcdraw]{xcolor} 
\usepackage{graphicx,graphics}
\usepackage{caption}
\usepackage{subcaption}
\usepackage{multirow}
\usepackage{wrapfig}
\usepackage{caption}
\usepackage{listings}
\usepackage[ruled,vlined]{algorithm2e}
\usepackage{pifont}

\def\gconv{\texttt{Graph Conv}}
\def\valign{\texttt{Vert Align}}

\newcommand{\mypar}[1]{\vspace{0.5mm}\noindent\textbf{#1}}
\newcolumntype{L}[1]{>{\raggedright\arraybackslash}p{#1}}
\newcolumntype{C}[1]{>{\centering\arraybackslash}p{#1}}
\newcommand{\bul}{\ding{55}}
\newcommand\FramedBox[3]{%
  \setlength\fboxsep{0pt}
  \setlength{\fboxrule}{0pt}
  \fbox{\parbox[b][#1][c]{#2}{\centering #3}}}

\title{Accelerating 3D Deep Learning with PyTorch3D}

\author{
  Nikhila Ravi \hspace{2mm}
  Jeremy Reizenstein \hspace{2mm}
  David Novotny \hspace{2mm}
  Taylor Gordon \\*[1mm]
  \textbf{Wan-Yen Lo} \hspace{2mm}
  \textbf{Justin Johnson}$^*$ \hspace{2mm}
  \textbf{Georgia Gkioxari}\thanks{equal contribution} \\*[2mm]
  Facebook AI Research
}

\begin{document}

\maketitle

\vspace{-5mm}
\begin{abstract}
Deep learning has significantly improved 2D image recognition. Extending into 3D may advance many new applications including autonomous vehicles, virtual and augmented reality, authoring 3D content, and even improving 2D recognition. However despite growing interest, 3D deep learning remains relatively underexplored. We believe that some of this disparity is due to the engineering challenges involved in 3D deep learning, such as efficiently processing heterogeneous data and reframing graphics operations to be differentiable. We address these challenges by introducing PyTorch3D, a library of modular, efficient, and differentiable operators for 3D deep learning. It includes a fast, modular differentiable renderer for meshes and point clouds, enabling analysis-by-synthesis approaches. Compared with other differentiable renderers, PyTorch3D is more modular and efficient, allowing users to more easily extend it while also gracefully scaling to large meshes and images. We compare the PyTorch3D operators and renderer with other implementations and demonstrate significant speed and memory improvements. We also use PyTorch3D to improve the state-of-the-art for unsupervised 3D mesh and point cloud prediction from 2D images on ShapeNet. PyTorch3D is open-source and we hope it will help accelerate research in 3D deep learning.
\end{abstract}

\section{Introduction}
\label{sec:intro}

Over the past decade, deep learning has significantly advanced the ability of AI systems to process 2D image data.
We can now build high-performing systems for tasks such as object~\cite{russakovsky2015imagenet,krizhevsky2012imagenet,simonyan2014very,he2016deep} and scene~\cite{yu2015lsun,zhou2017places} classification, object detection~\cite{ren2015faster}, semantic~\cite{long2015fully} and instance~\cite{he2017maskrcnn} segmentation, and human pose estimation~\cite{alp2018densepose}.
These systems can operate on complex image data and have been deployed in countless real-world settings.
Though sucessful, these methods suffer from a common shortcoming: they process 2D snapshots and ignore the true 3D nature of the world.

Extending deep learning into 3D can unlock many new applications.
Recognizing objects in 3D point clouds~\cite{qi2017pointnet,qi2017pointnet++} can enhance the sensing abilities of autonomous vehicles, or enable new augmented reality experiences.
Predicting depth~\cite{eigen2014depth,eigen2015predicting}, or 3D shape~\cite{choy2016r2n2,fan2017point,wang2018pixel2mesh,novotny2019c3dpo} can lift 2D images into 3D.
Generative models~\cite{wu2016learning,yang2019pointflow,nash2020polygen} might one day aid artists in authoring 3D content.
Image-based tasks like view synthesis can be improved with 3D representations given only 2D supervision~\cite{sitzmann2019scene,wiles2020synsin,mildenhall2020nerf}.
Despite growing interest, 3D deep learning remains relatively underexplored.

We believe that some of this disparity is due to the significant engineering challenges involved in 3D deep learning.
One such challenge is \emph{heterogeneous data}.
2D images are almost universally represented by regular pixel grids.
In contrast, 3D data are stored in a variety of structured formats including voxel grids~\cite{choy2016r2n2,tatarchenko2017octree}, point clouds~\cite{qi2017pointnet,fan2017point}, and meshes~\cite{wang2018pixel2mesh,gkioxari2019mesh} which can exhibit per-element heterogeneity.
For example, meshes may differ in their number of vertices and faces, and their topology.
Such heterogeneity makes it difficult to efficiently implement batched operations on 3D data using the tensor-centric primitives provided by standard deep learning toolkits like PyTorch~\cite{paszke2019pytorch} and Tensorflow~\cite{abadi2016tensorflow}.
A second key challenge is \emph{differentiability}.
The computer graphics community has developed many methods for efficiently processing 3D data.
However, to be embedded in deep learning pipelines, each operation must be revisited to also efficiently compute gradients.
Some operations, such as camera transformations, admit gradients trivially via automatic differentiation.
Others, such as mesh rendering, must be reformulated via differentiable relaxation~\cite{loper2014opendr,kato2018neural,liu2019soft,chen2019learning}.

We address these challenges by introducing PyTorch3D, a library of modular, efficient, and well-tested operators for 3D deep learning built on PyTorch~\cite{paszke2019pytorch}.
All operators are fast and differentiable, and many are implemented with custom CUDA kernels to improve efficiency and minimize memory usage.
We provide reusable data structures for managing batches of point cloud and meshes, allowing all PyTorch3D operators to support batches of heterogeneous data.

One key feature of PyTorch3D is a modular and efficient differentiable rendering engine for meshes and point clouds.
Differentiable rendering projects 3D data to 2D images, enabling \emph{analysis-by-synthesis} \cite{grenander} and \emph{inverse rendering} \cite{marschner1998inverse,patow2003survey} approaches where 3D predictions can be made using only image-level supervision~\cite{loper2014opendr}. 
Compared to other recent differentiable renderers~\cite{kato2018neural,liu2019soft}, ours is both more modular and more scalable.
We achieve modularity by decomposing the rendering pipeline into stages (rasterization, lighting, shading, blending) which can easily be replaced with new user-defined components, allowing users to adapt our renderer to their needs.
We improve efficiency via two-stage rasterization, and by limiting the number of primitives that influence each pixel.

We compare the optimized PyTorch3D operators with na\"ive PyTorch implementations and with those provided in other open-source packages, demonstrating improvements in speed and memory usage by up to 10$\times$.
We also showcase the flexibility of PyTorch3D by experimenting on the task of unsupervised shape prediction on the ShapeNet~\cite{shapenet} dataset.
With our modular differentiable renderer and efficient 3D operators we improve the state-of-the-art for unsupervised 3D mesh and point cloud prediction from 2D images while maintaining high computational throughput.

PyTorch3D is open-source\footnote{\url{https://pytorch3d.org/}} and will evolve over time.
We hope that it will be a valuable tool to the community and help accelerate research in 3D deep learning.

\vspace{-3mm}
\section{Related Work}
\label{sec:related}
\vspace{-2mm}
\mypar{3D deep learning libraries.}
There are a number of toolkits for 3D deep learning. \cite{FeyLenssen2019} focuses on learning on graphs, \cite{TensorflowGraphicsIO2019} provides differentiable graphics operators, \cite{jatavallabhula2019kaolin} collects commonly used 3D functions. However, they do not provide support for heterogeneous batching of 3D data, crucial for large-scale learning, or modularity for differentiable rendering, crucial for exploration. PyTorch3D introduces data structures that support batches of 3D data with varying sizes and topologies. This key abstraction allows our 3D operators, including rendering, to operate on large heterogeneous batches.

\mypar{Differentiable renderers.}
OpenDR~\cite{loper2014opendr} and NMR~\cite{kato2018neural} perform traditional rasterization in the forward pass and compute approximate gradients in the backward pass.
More recently, SoftRas~\cite{liu2019soft} and DIB-R~\cite{chen2019learning} propose differentiable renderers by viewing rasterization as a probabilistic process where each pixel's color depends on multiple mesh faces.
Differentiable ray tracing methods, such as Redner~\cite{li2018differentiable} and Mitsuba2~\cite{nimier2019mitsuba}, give more photorealistic images at the expense of increased compute.

Differentiable point cloud rendering is explored in~\cite{insafutdinov2018unsupervised} which uses ray termination probabilities and stores points in a voxel grid which limits resolution. DSS~\cite{yifan2019differentiable} renders each point as a disk. SynSin~\cite{wiles2020synsin} also splats a per point sphere from a soft z-buffer of fixed length. Most recently, Pulsar~\cite{lassner2020fast} uses an unlimited z-buffer for rendering but uses the first few points for gradient propagation.

Differentiable rendering is an active research area.
PyTorch3D introduces a modular renderer, inspired by~\cite{liu2019soft}, by redesigning and exposing intermediates computed during rasterization.
Unlike other differentiable renderers, users can easily customize the rendering pipeline with PyTorch shaders.

\mypar{3D shape prediction.}
In Section~\ref{sec:exp} we experiment with unsupervised 3D shape prediction using the differentiable silhouette and textured renderers for meshes and point clouds in PyTorch3D. There is a vast line of work on 3D shape prediction, including two-view and multiview methods~\cite{scharstein2002taxonomy,hartley2003multiple}, model-based approaches~\cite{fischler1973representation,brooks1979acronym,brooks1983model,lowe1991fitting,blanz2003face,loper2015smpl}, and recent 
supervised deep methods that predict voxels~\cite{choy2016r2n2,tatarchenko2017octree}, meshes~\cite{groueix2018papier,wang2018pixel2mesh,smith2019geometrics,gkioxari2019mesh}, point clouds~\cite{fan2017point} and implicit functions~\cite{mescheder2019occupancy,park2019deepsdf,saito2019pifu}. 
Differentiable renderers allow for unsupervised shape prediction via 2D re-projection losses~\cite{tulsiani2017multi,kato2018neural,kanazawa2018learning,lsiTulsiani18,liu2019soft}.

\section{PyTorch3D: Functionality and Performance}
\label{sec:bench}

This section describes the core features of PyTorch3D.
For a 3D deep learning library to be effective, 3D operators need to be efficient when handling complex 3D data.
We benchmark the speed and memory usage of key PyTorch3D operators, comparing to pure PyTorch and existing open-source implementations.
We show that PyTorch3D achieves speedups up to $10\times$.
 
\mypar{3D data structures.}
Working with minibatches of data is crucial in deep learning both for stable optimization and computational efficiency.
However operating on batches of 3D meshes and point clouds is challenging due to \emph{heterogeneity}: meshes may have varying numbers of vertices and faces, and point clouds may have varying numbers of points.
To overcome this challenge, PyTorch3D provides data structures to manage batches of meshes and point clouds which allow conversion between different tensor-based representations (\emph{list}, \emph{packed}, \emph{padded}) needed for various operations.
 
\mypar{Implementation Details.}
We benchmark with meshes from ShapeNetCoreV1~\cite{shapenet} using \emph{homogeneous} and \emph{heterogeneous} batches.
We form point clouds by sampling uniformly from mesh surfaces.
All results are averaged over 5 random batches and 10 runs per batch, and are run on a V100 GPU.

\subsection{3D operators}

We report time and memory usage for a representative set of popular 3D operators, namely \emph{Chamfer loss}, \emph{graph convolution} and \emph{K nearest neighbors}. Other 3D operators in PyTorch3D follow similar trends. We compare to PyTorch and state-of-the-art open-source libraries.

\begin{figure}
\hspace{-0.1in}
\begin{subfigure}{0.26\textwidth}
\includegraphics[width=1.02\columnwidth, height=1.1\columnwidth]{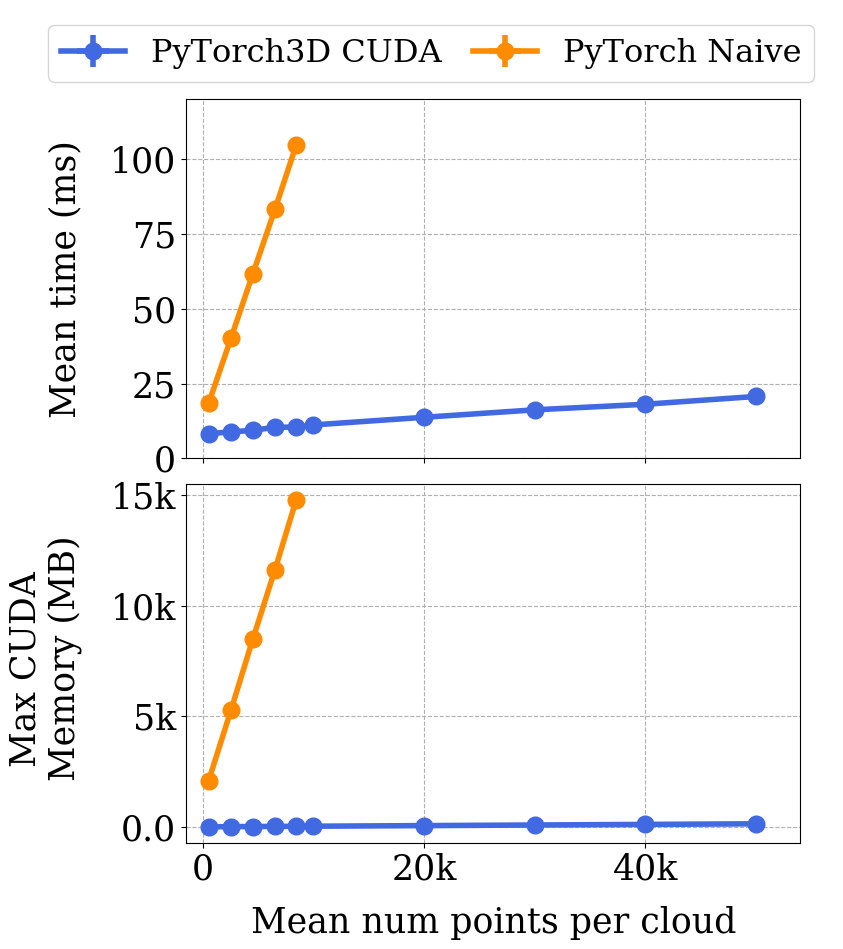}
\caption{Chamfer}
\label{fig:bench_chamfer}
\end{subfigure}
\hspace{-0.1in}
\begin{subfigure}{0.26\textwidth}
  \vspace{-0.3mm}
  \includegraphics[width=0.97\columnwidth, height=1.1\columnwidth]{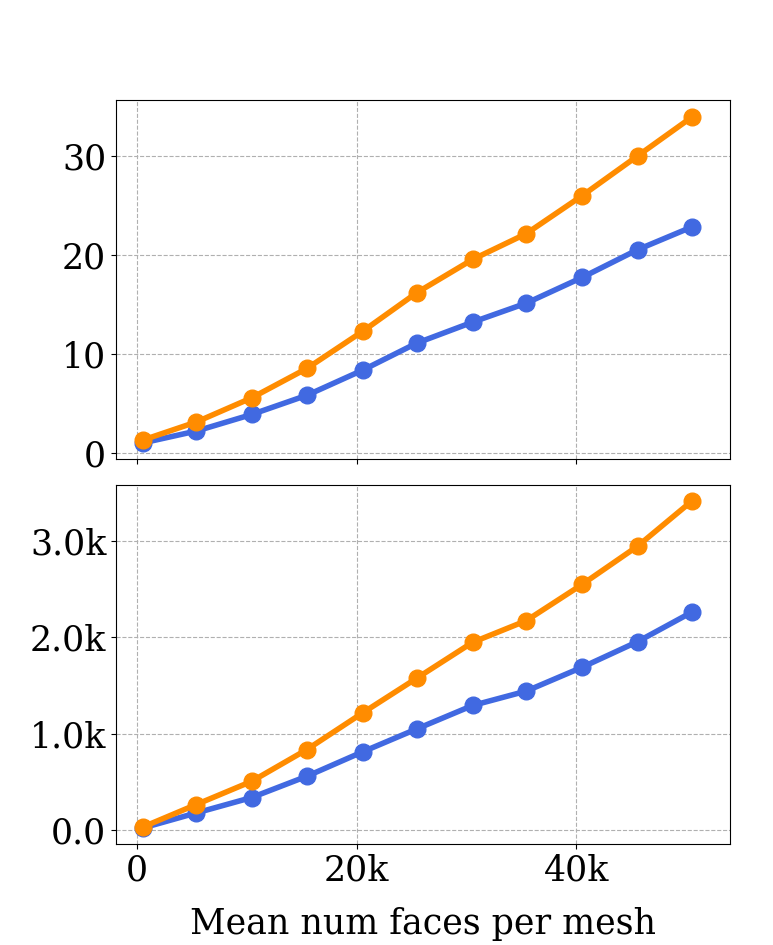}
  \caption{Graph Conv}
  \label{fig:bench_gconv}
\end{subfigure}
\hspace{-0.1in}
\begin{subfigure}{0.26\textwidth}
  \includegraphics[width=0.93\columnwidth, height=1.1\columnwidth]{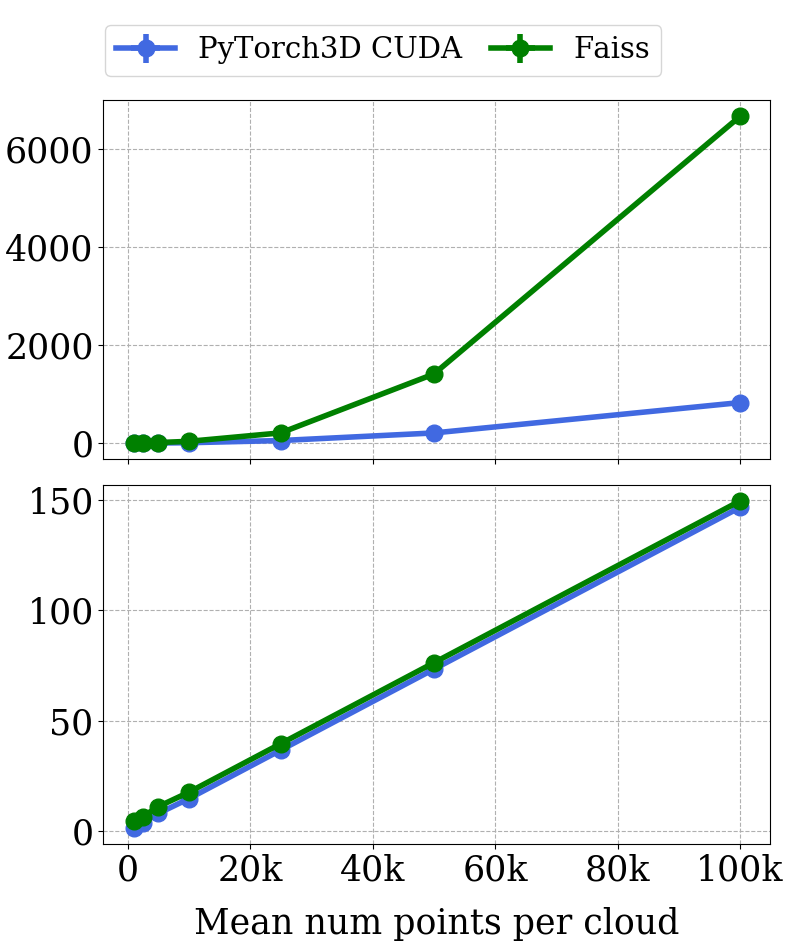}
  \caption{KNN (varying $|P|$)}
  \label{fig:bench_knn_p}
\end{subfigure}
\hspace{-0.12in}
\begin{subfigure}{0.26\textwidth}
  \includegraphics[width=0.93\columnwidth, height=1.1\columnwidth]{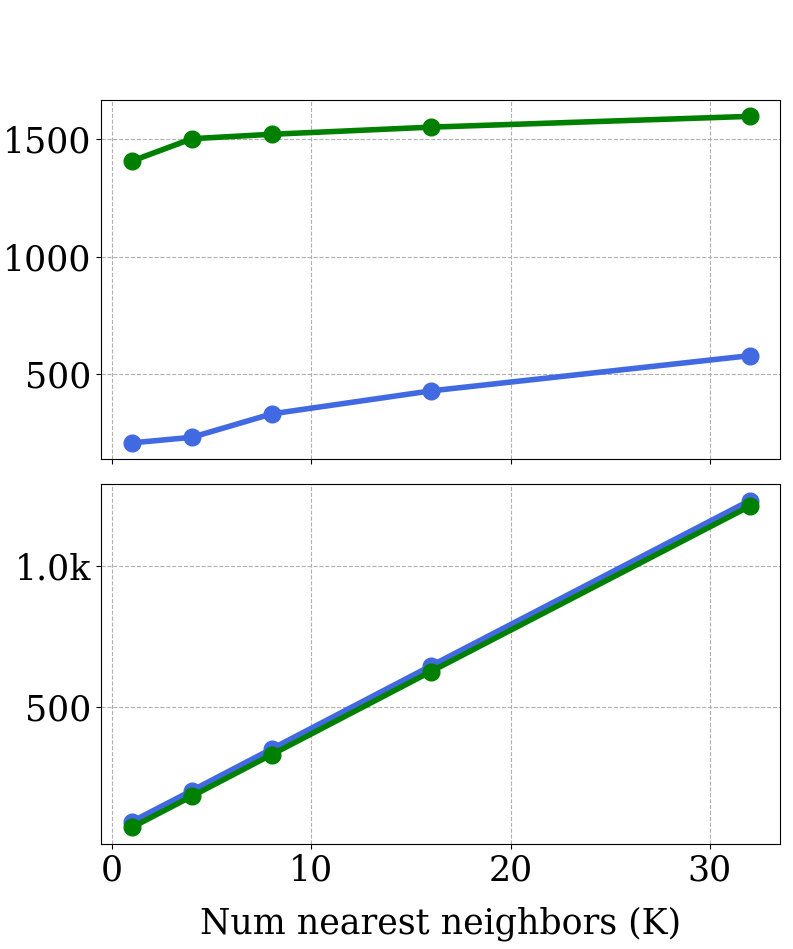}
  \caption{KNN (varying K)}
  \label{fig:bench_knn_k}
\end{subfigure}
\vspace{-2mm}
\caption{
  Benchmarks for our 3D operators with batch size 32.
  (a) $\mathcal{L}_{\textrm{cham}}(P, Q)$ for point clouds with $|P|=1000$ and heterogeneous and varying $|Q|$.
  (b) Graph convolution on heterogeneous mesh batches with 128-dimensional features.
  (c,d) Our KNN vs Faiss~\cite{johnson2017faiss} between homogeneous batches of 3D point clouds $P$ and $Q$ with $|P|=|Q|$.
  In (c) $K=1$, and in (d) $|P|=|Q|=50$k; in both our memory usage matches \cite{johnson2017faiss}.
  (a) and (b) are forward and backward; (c) and (d) are forward only.
}
\label{fig:benchmark_3D_ops}
\vspace{-3mm}
\end{figure}

\mypar{Chamfer loss}
is a common metric that quantifies agreement between point clouds $P$ and $Q$. Formally,
\begin{equation}
  \small
  \mathcal{L}_{\textrm{cham}}(P, Q) =
    |P|^{-1} \hspace{-4mm} \sum_{(p,q)\in\Lambda_{P,Q}}\hspace{-3mm}\|p-q\|^2
  + |Q|^{-1} \hspace{-4mm} \sum_{(q,p)\in\Lambda_{Q,P}}\hspace{-3mm}\|q-p\|^2
\end{equation}
where $\Lambda_{P,Q} = \{(p, \arg\min_q\|p-q\|) : p\in P\}$ is the set of pairs $(p, q)$ such that $q \in Q$ is the nearest neighbor of $p \in P$.
A (homogeneously) batched implementation is straightforward in PyTorch, but is inefficient since it requires forming a pairwise distance matrix with $B\times|P|\times|Q|$ elements (where $B$ is the batch size).
PyTorch3D avoids this inefficiency (and supports heterogeneity) by using our efficient KNN to compute neighbors.
Figure~\ref{fig:bench_chamfer} compares ours against the na\"{i}ve approach with $B=32$, $|P|=1000$, and varying $|Q|$.
The na\"{i}ve approach runs out of memory for $|Q|>10$k, while ours scales to large point clouds and reduces time and memory use by more than $12\times$. 
 
\mypar{Graph convolution}~\cite{kipf2017semi} is commonly used for processing 3D meshes~\cite{wang2018pixel2mesh,gkioxari2019mesh}.
Given feature vectors $f_v$ for each vertex $v$, it computes new features $f'_v=W_0f_v + \sum_{u\in\mathcal{N}(v)}W_1f_u$ where $\mathcal{N}(v)$ are the neighbors of $v$ in the mesh and $W_0,W_1$ are learned weight matrices.
PyTorch3D implements graph convolution via a fused CUDA kernel for \texttt{gather}$+$\texttt{scatter\_add}.
Figure~\ref{fig:bench_gconv} shows that this improves speed and memory use by up to $30\%$ compared against a pure PyTorch implementation.

\mypar{K Nearest Neighbors}
for $D$-dimensional points are used in Chamfer loss, normal estimation, and other point cloud operations.
We implement exact KNN with custom CUDA kernels that natively handle heterogeneous batches.
Our implementation is tuned for $D\leq4$ and $K\leq32$, and uses template metaprogramming to individually optimize each $(D, K)$ pair.
We compare against Faiss~\cite{johnson2017faiss}, a fast GPU library for KNN that targets a different portion of the design space: it does not handle batching, is optimized for high-dimensional descriptors ($D\approx128$), and scales to billions of points.
Figures~\ref{fig:bench_knn_p} and \ref{fig:bench_knn_k} show that we outperform Faiss by up to $5\times$ for batched 3D problems.
 
\begin{figure}
\hspace{-0.4in}
\begin{subfigure}{0.58\textwidth}
\includegraphics[width=0.99\columnwidth,height=0.584\columnwidth]{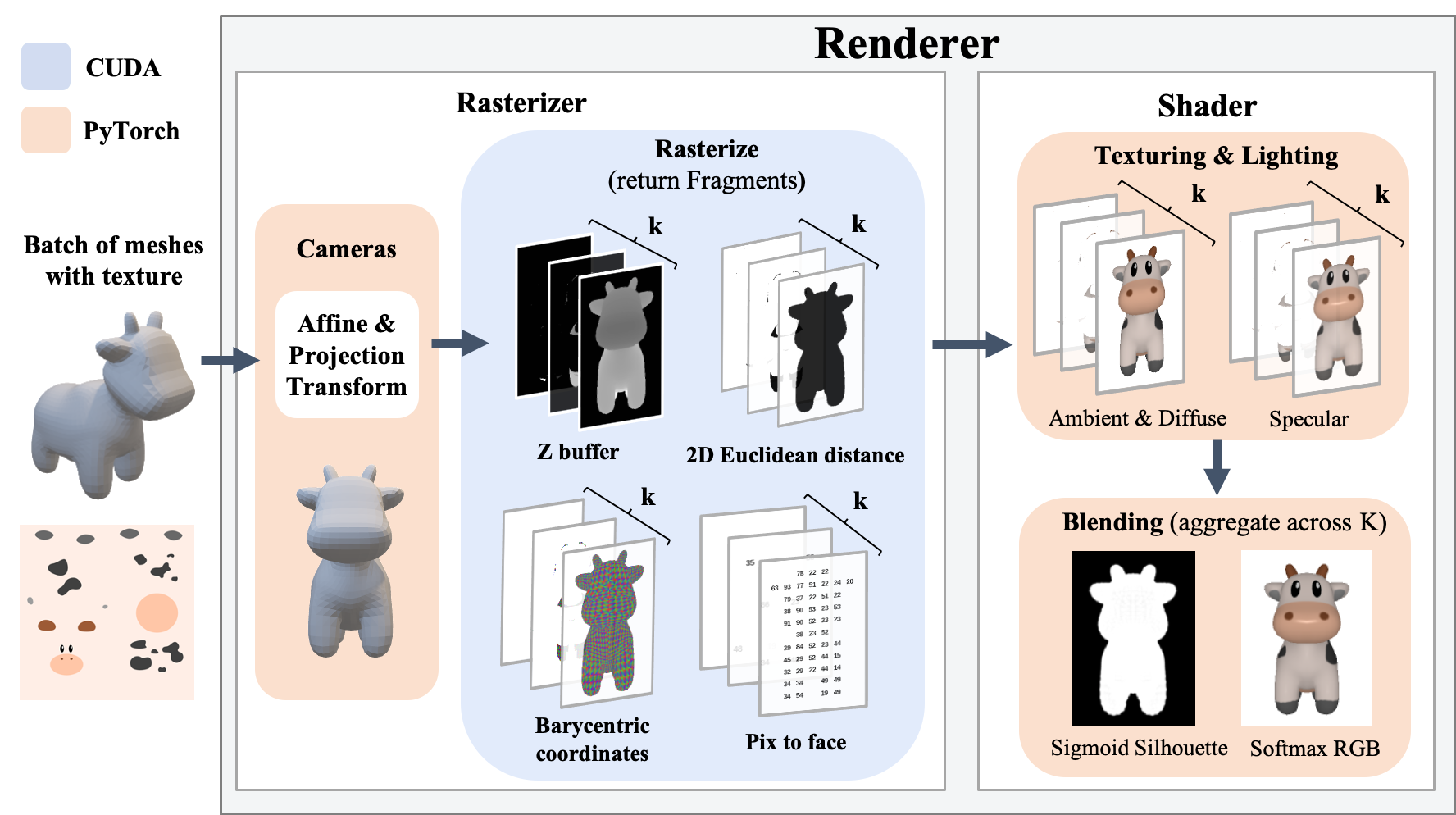}
\caption{PyTorch3D rendering pipeline}
\label{fig:renderer_overview}
\end{subfigure}
\hfill
\begin{subfigure}{0.51\textwidth}
\vspace{-0.8mm} 
\includegraphics[width=0.99\columnwidth]{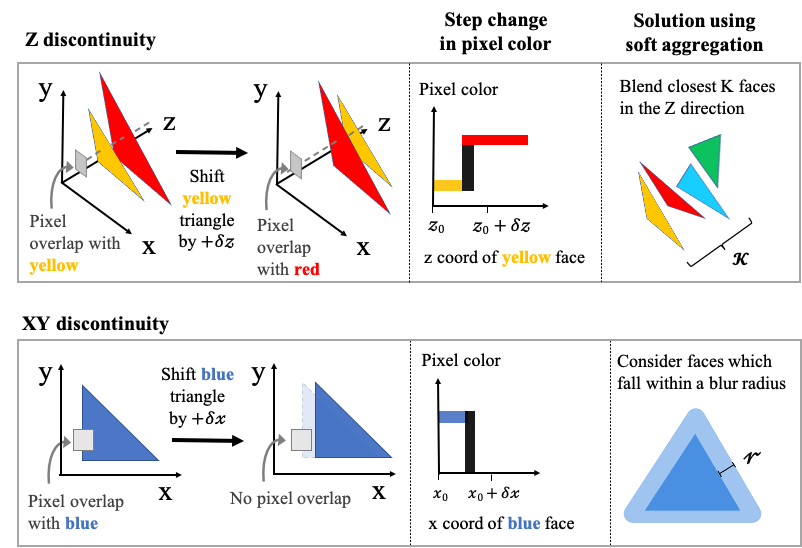}
\vspace{-0.6mm} 
\caption{Problems with differentiability in rendering}
\label{fig:renderer_diff}
\end{subfigure}
\vspace{-1mm}
\caption{(a) The modular rendering pipeline in PyTorch3D and (b) The $z$- \& $xy$-discontinuities in traditional rasterization and the soft formulations~\cite{liu2019soft} which enable differentiability.}
 \label{fig:renderer_diagram}
\vspace{-3mm}
\end{figure}

\subsection{Differentiable mesh renderer}
A \emph{renderer} inputs scene information (camera, geometry, materials, lights, textures) and outputs an image.
A \emph{differentiable} renderer can also propagate gradients backward from rendered images to scene information~\cite{loper2014opendr}, allowing rendering to be embedded into deep learning pipelines~\cite{kato2018neural,liu2019soft}.

PyTorch3D includes a differentiable renderer that operates on heterogeneous batches of triangle meshes.
Our renderer follows three core design principles:
\emph{differentiability}, meaning that it computes gradients with respect to all inputs; \emph{efficiency}, meaning that it runs quickly and scales to large meshes and images; and \emph{modularity}, meaning that users can easily replace components of the renderer to customize its functionality to their use case and experiment with alternate formulations. 

As shown in Figure~\ref{fig:renderer_overview}, our renderer has two main components: the \emph{rasterizer} selects the faces affecting each pixel, and the \emph{shader} computes pixel colors.
Through careful design of these components, 
we improve efficiency and modularity compared to prior differentiable renderers~\cite{kato2018neural,liu2019soft,chen2019learning}.

\mypar{Rasterizer.}
The rasterizer first uses a \emph{camera} to transform meshes from world to view coordinates.
Cameras are Python objects and compute gradients via autograd; this aids modularity, as users can easily implement new camera models other than our provided orthographic and perspective cameras.

Next, the core rasterization algorithm finds triangles that intersect each pixel.
In traditional rasterization, each pixel is influenced only by its nearest face along the $z$-axis.
As shown in Figure~\ref{fig:renderer_diff}, this can cause step changes in pixel color as faces move along the $z$-axis (due to occlusion) and in the $xy$-plane (due to face boundaries).
Following \cite{liu2019soft} we soften these nondifferentiabilities by blending the influence of multiple faces for each pixel, and decaying a face's influence toward its boundary.

Our rasterizer departs from \cite{liu2019soft} in three ways to improve efficiency and modularity.
First, in \cite{liu2019soft}, pixels are influenced by every face they intersect in the $xy$-plane; in contrast we constrain pixels to be influenced by only the nearest $K$ faces along the $z$-axis, computed using per-pixel priority queues.
Similar to traditional $z$-buffering, this lets us quickly discard many faces for each pixel, improving efficiency.
We show in Section~\ref{sec:exp} that this modification does not harm downstream task performance.
Second, \cite{liu2019soft} na\"ively compares each pixel with each face.
We improve efficiency using a two-pass approach similar to \cite{laine2011high},
first working on image tiles to eliminate faces before moving to pixels.
Third, \cite{liu2019soft} fuses rasterization and shading into a monolithic CUDA kernel.
We decouple these, and as shown in Figure~\ref{fig:renderer_overview} our rasterizer returns \emph{Fragment data} about the $K$ nearest faces to each pixel: face ID, barycentric coordinates of the pixel in the face, and (signed) pixel-to-face distances along the $z$-axis and in the $xy$-plane.
This allows shaders to be implemented separately from the rasterizer, significantly improving modularity.
This change also improves efficiency, as cached Fragment data can be used to avoid costly recomputation of face-pixel intersections in the backward pass.

\begin{figure}
\hspace{-0.23in}
\begin{subfigure}{0.41\textwidth}
\includegraphics[width=0.97\columnwidth, height=0.8\columnwidth]{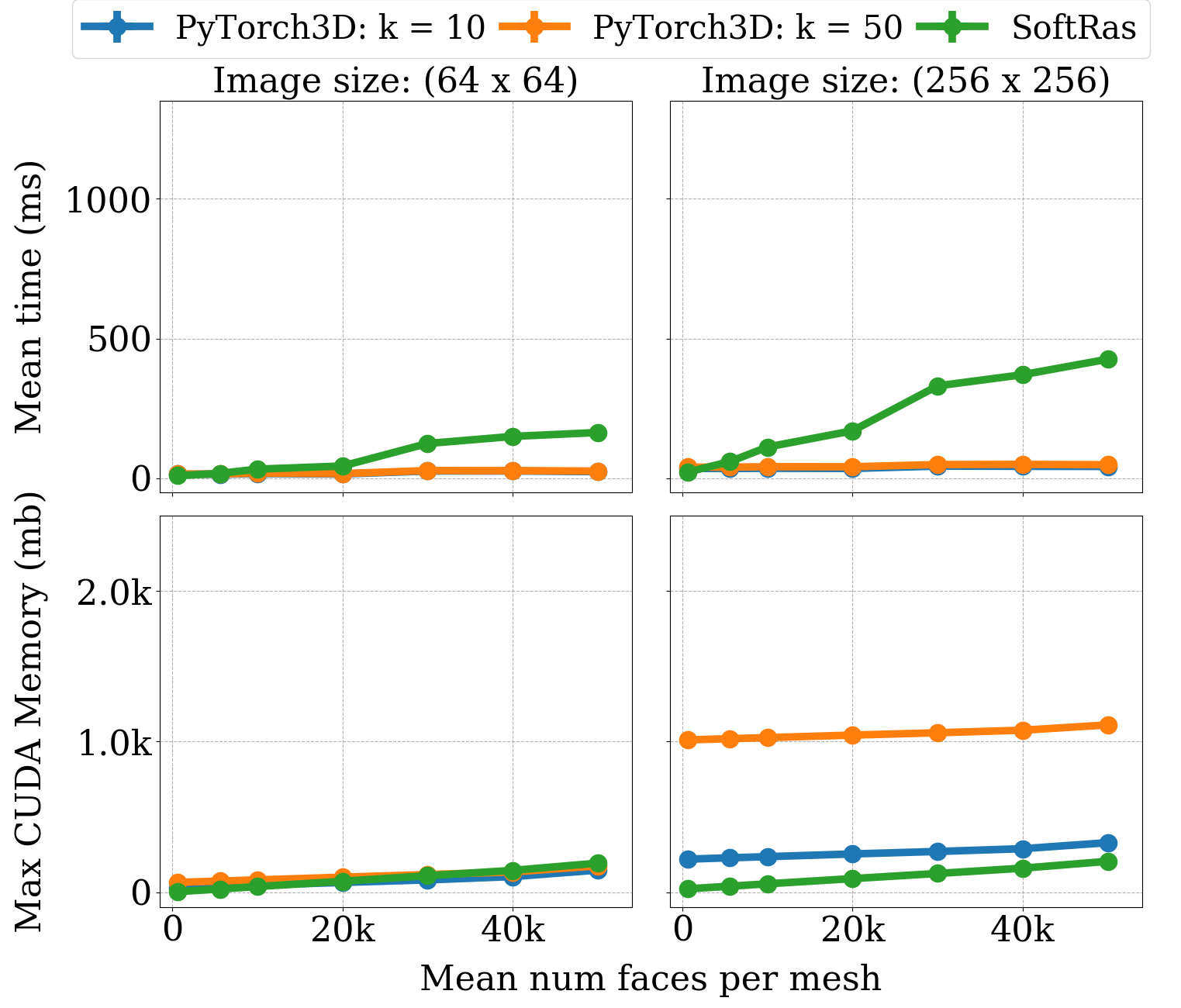}
\caption{\textbf{Silhouette} homogeneous}
\label{fig:silh_render_hom}
\end{subfigure}
\hspace{-0.15in}
\begin{subfigure}{0.41\textwidth}
\includegraphics[width=0.93\columnwidth, height=0.8\columnwidth]{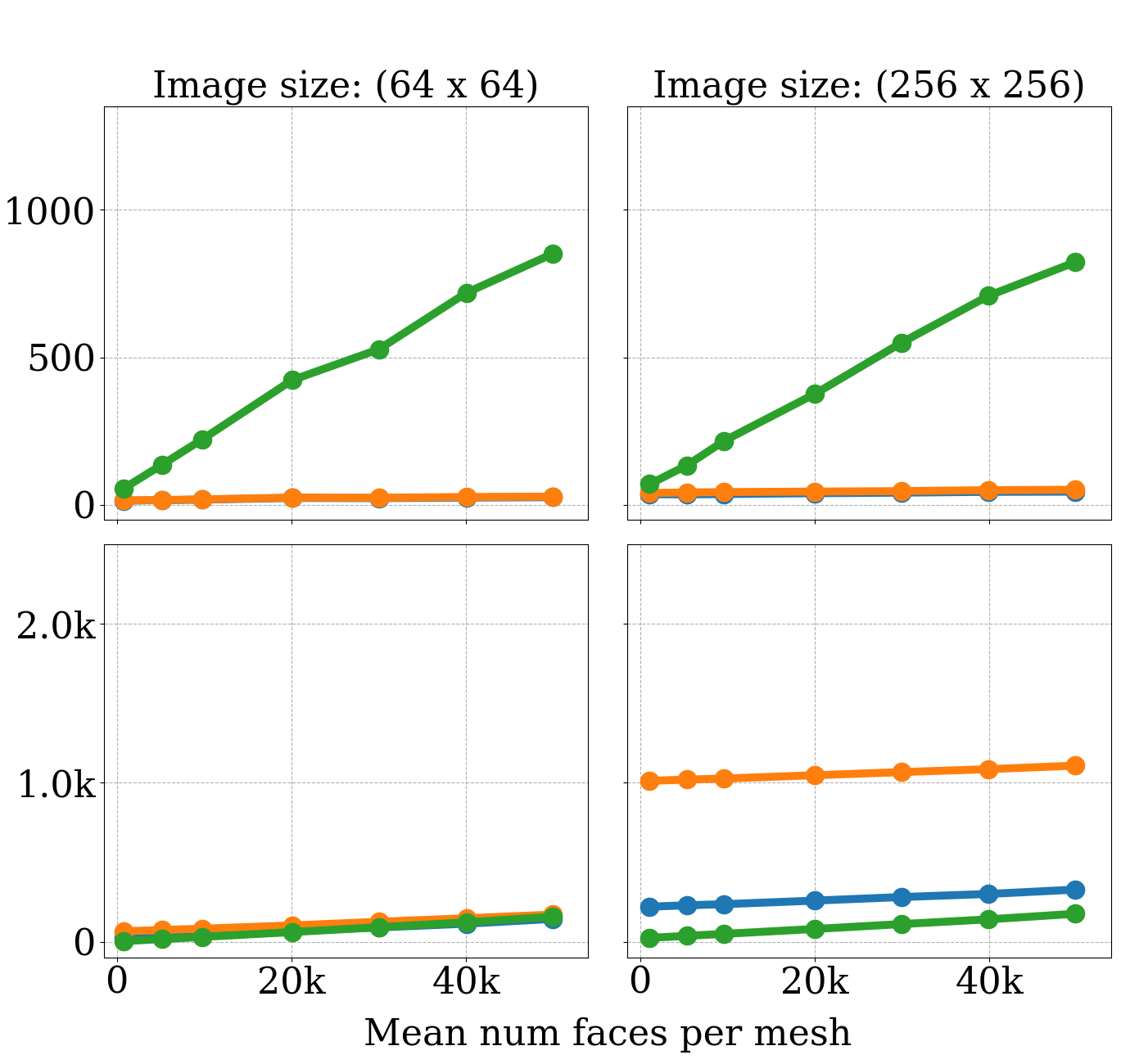}
\caption{\textbf{Silhouette} heterogeneous}
\label{fig:silh_render_het}
\end{subfigure}
\hspace{-0.2in}
\begin{subfigure}{0.26\textwidth}
\includegraphics[width=0.8\columnwidth, height=1.26\columnwidth]{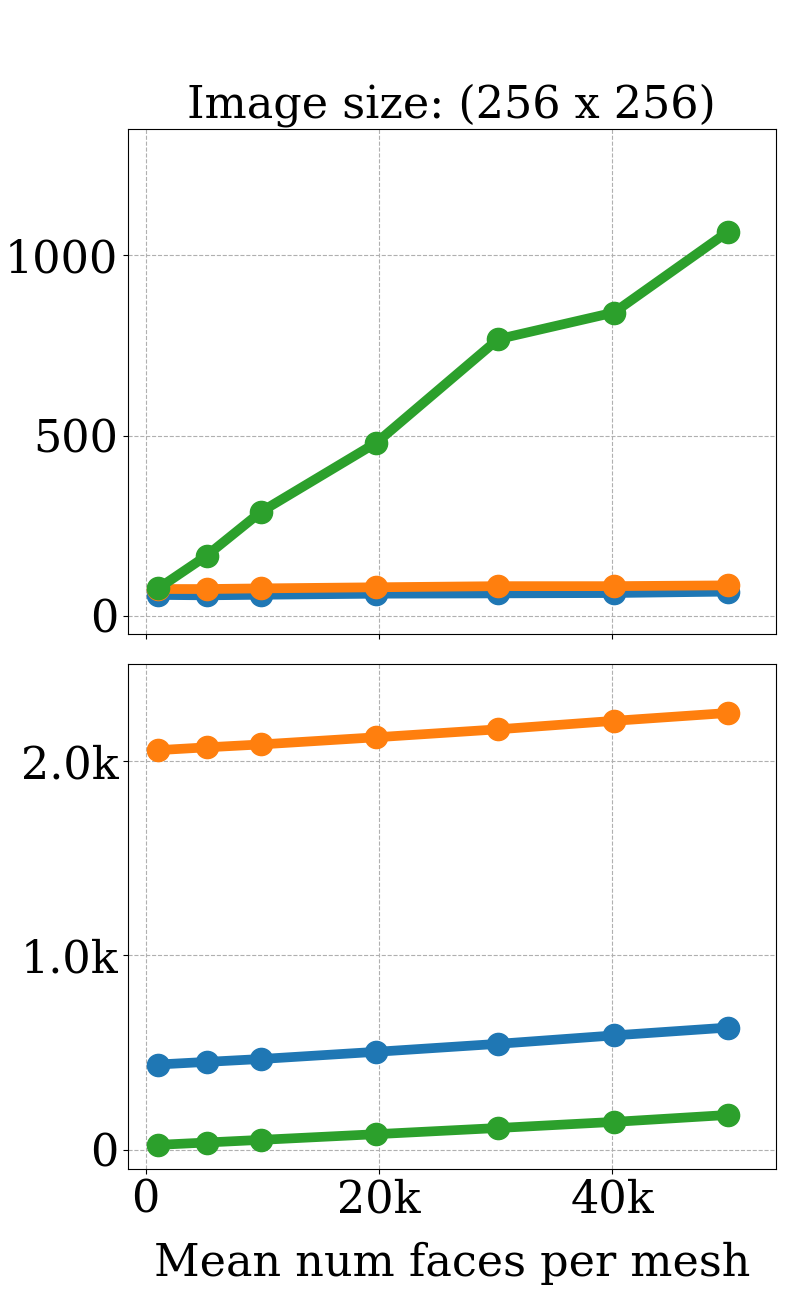}
\caption{\textbf{Texture} heterogeneous}
\label{fig:text_render}
\end{subfigure}
\vspace{-2mm}
\caption{Benchmarks for \emph{silhouette} and \emph{textured} rendering for PyTorch3D and SoftRas~\cite{liu2019soft}. We use a batch size of 8, two image sizes (64 \& 256) and two values for the number of faces per pixel K {10 \& 50} (for PyTorch3D only). All benchmarks are for forward and backward.}
\label{fig:benchmark_mesh_render}
 \vspace{-5mm}
\end{figure}

\mypar{Shaders}
consume the Fragment data produced by the rasterizer, and compute pixel values of the rendered image.
They typically involve two stages: first computing $K$ values for the pixel (one for each face identified by the Fragment data), then \emph{blending} them to give a final pixel value.

\begin{wraptable}{r}{0.6\textwidth}
\vspace{-4mm}
	\begin{minipage}{0.6\textwidth}
	\begin{algorithm}[H]
	prob = (-- dists / sigma).sigmoid() \\ 
	alpha = 1 -- (1 -- prob).prod(dim=--1)
	\caption{Silhouette~blending}
	\label{algo:silh2}
	\end{algorithm}
	\end{minipage}
	\begin{minipage}{0.6\textwidth}
	\begin{algorithm}[H]
	prob = (-- dists / sigma).sigmoid() \\ 
	zinv = (zfar -- zbuf) / (zfar -- znear)\\
	zinv\_max = torch.max(zinv, dim=--1).values \\
	weights = prob * ((zinv -- zinv\_max) / gamma)).exp() \\	
	weights = weights / weights.sum(dim=--1) \\
	image = (weights * top\_k\_colors\_per\_pixel).sum(dim=--2)
	\caption{Softmax blending}
	\label{algo:soft}
	\end{algorithm}
	\end{minipage}
\vspace{-4mm}
\end{wraptable}
Shaders are Python objects, and Fragment data are stored in PyTorch tensors.
Shaders can thus work with $Fragment$ data using standard PyTorch operators, and compute gradients via autograd.
This design is highly modular, as users can easily implement new shaders to customize the renderer.
For example, Algorithm~\ref{algo:silh2} implements the silhouette renderer from \cite{liu2019soft} using a two-line shader:
\verb|dists| is a tensor of shape $B\times H\times W\times K$ giving signed distances in the $xy$-plane from each pixel to its $K$ nearest faces (part of Fragment data), and \verb|sigma| is a hyperparameter.
This is simpler than \cite{liu2019soft} where silhouette rendering is one path in a monolithic CUDA kernel and gradients are manually computed.  Similarly, Algorithm~~\ref{algo:soft} implements the softmax blending algorithm from~\cite{liu2019soft} for textured rendering. $dists$, $zbuf$ are part of the Fragment data, and $top\_k\_colors\_per\_pixel$ is the output from the shader.  $zfar$, $znear$, $sigma$, $gamma$ are hyper-parameters defined by the user. 

Shaders can implement complex effects using the Fragment data from the rasterizer.
Face IDs can be used to fetch per-face data like normals, colors, or texture coordinates; barycentric coordinates can be used to interpolate data over the face; $xy$ and $z$ distances can be used to blend the influence of faces in different ways.
Crucially, all texturing, lighting, and blending logic can be written using PyTorch operators, and differentiated using autograd.
We provide a variety of shaders implementing silhouette rendering, flat, Gouraud~\cite{gouraud1971continuous}, and Phong~\cite{phong1975illumination} shading with per-vertex colors or texture coordinates, and which blend colors using hard assignment (similar to \cite{kato2018neural}) or softmax blending (like \cite{liu2019soft}).

\mypar{Performance.}
In Figure~\ref{fig:benchmark_mesh_render} we benchmark the speed and memory usage of our renderer against SoftRas~\cite{liu2019soft}.
We implement shaders to reproduce their silhouette rendering and textured mesh rendering using per-vertex textures and Gouraud shading.
Ours is significantly faster, especially for large meshes, higher-resolution images, and heterogeneous batches:
for textured rendering of heterogenous batches of meshes with mean 50k faces each at $256\times256$, our renderer is more than $4\times$ faster than \cite{liu2019soft}.
Our renderer uses more GPU memory than \cite{liu2019soft} since we explicitly store Fragment data.
However our absolute memory use ($\approx$ 2GB for texture at $256^2$)
is small compared to modern GPU capacity (32GB for V100);
we believe our improved modularity offsets our memory use.

\subsection{Differentiable point cloud renderer}
PyTorch3D also provides an efficient and modular point cloud renderer following the same design as the mesh renderer.
It is similarly factored into a \emph{rasterizer} that finds the $K$-nearest points to each pixel along the $z$-direction,
and \emph{shaders} written in PyTorch that consume fragment data from the rasterizer to compute pixel colors.
We provide shaders for silhouette and textured point cloud rendering, and users can easily implement custom shaders to customize the rendering pipeline.
Like the mesh renderer, the point cloud render natively supports heterogeneous batches of points.

Our point cloud renderer uses a similar strategy as our mesh renderer for overcoming the non-differentiabilities discussed in Figure~\ref{fig:renderer_diff}.
Each point is splatted to a circular region in screen-space whose opacity decreases away from the region's center.
The value of each pixel is computed by blending information for the $K$-nearest points in the $z$-axis whose splatted regions overlap the pixel.

In our experiments we consider two blending methods: \emph{Alpha}-compositing and \emph{Norm}alized weighted sums.
Suppose a pixel is overlapped by the splats from $K$ points with opacities $\alpha_1,\ldots,\alpha_K\in[0,1]$ sorted in increasing $z$-order, and the points are associated with feature vectors $f_1,\ldots,f_K\in\mathbb{R}^D$.
Features might be boolean (for silhouette rendering), RGB colors (for textured rendering), or neural features~\cite{insafutdinov2018unsupervised,wiles2020synsin}.
The blending methods compute features $f_{Alpha},f_{Norm}\in\mathbb{R}^D$ for the pixel:
\begin{equation}
  f_{Alpha} = \sum_{i=1}^K\left(\alpha_i\prod_{j=1}^{i-1}(1-\alpha_j)\right)f_i
  \hspace{2pc}
  f_{Norm} = \left(\sum_{i=1}^K\alpha_i f_i\right)/\left(\sum_{i=1}^K\alpha_i\right).
\end{equation}
\emph{Alpha}-compositing uses the depth ordering of points so that nearer points contribute more, while \emph{Norm} ignores the depth order.
Both blending functions are differentiable and can propagate gradients from pixel features backward to both point features and opacities.
They can be implemented with a few lines of PyTorch code similar to Algorithms~\ref{algo:silh2} and \ref{algo:soft}.

\begin{figure}[t]
\centering
\begin{subfigure}{0.39\textwidth}
\centering
\includegraphics[height=4.9cm]{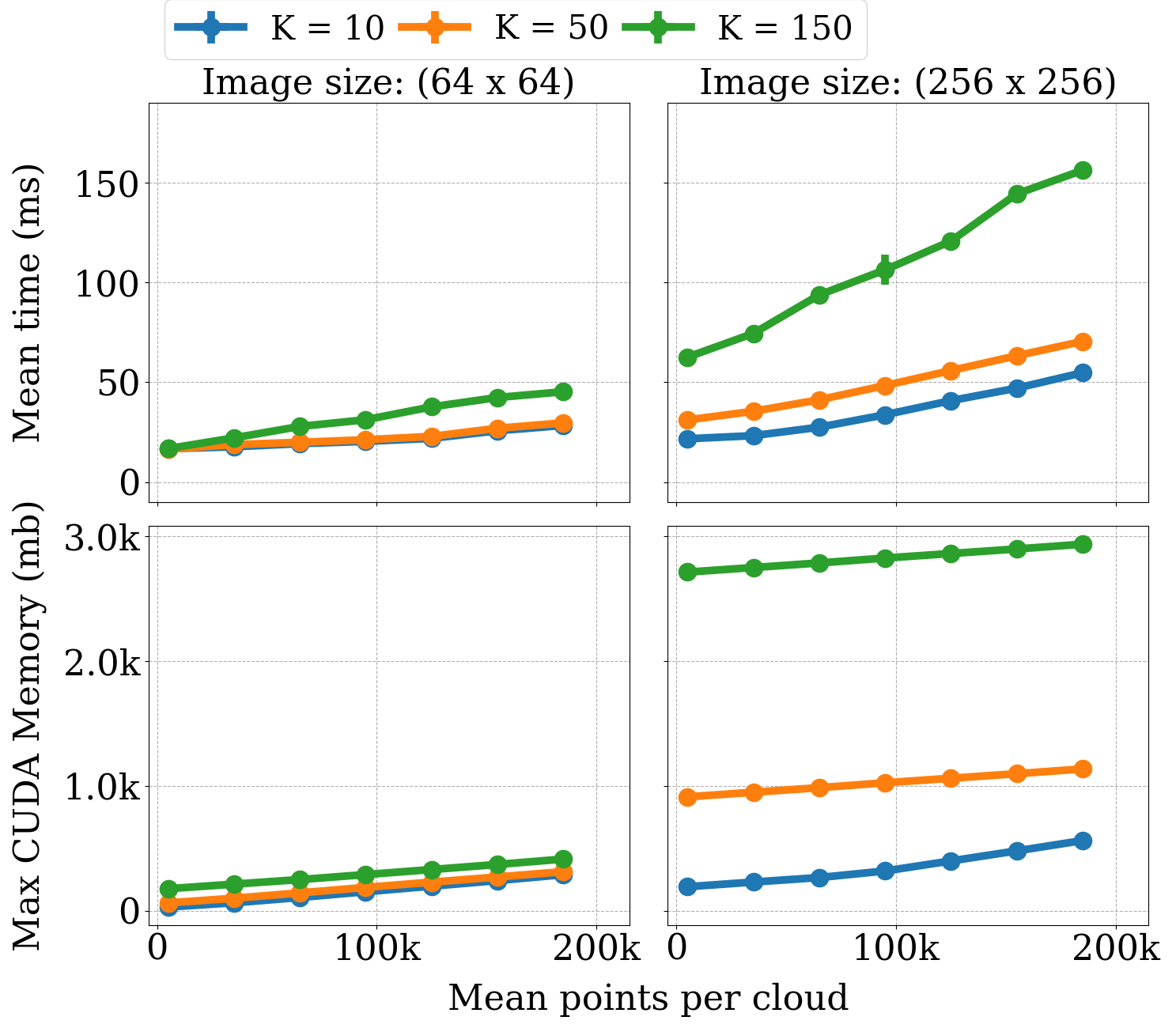}
\caption{\textbf{Alpha} homogeneous}
\label{fig:alpha_hom}
\end{subfigure}
\begin{subfigure}{0.345\textwidth}
\centering
\includegraphics[height=4.9cm]{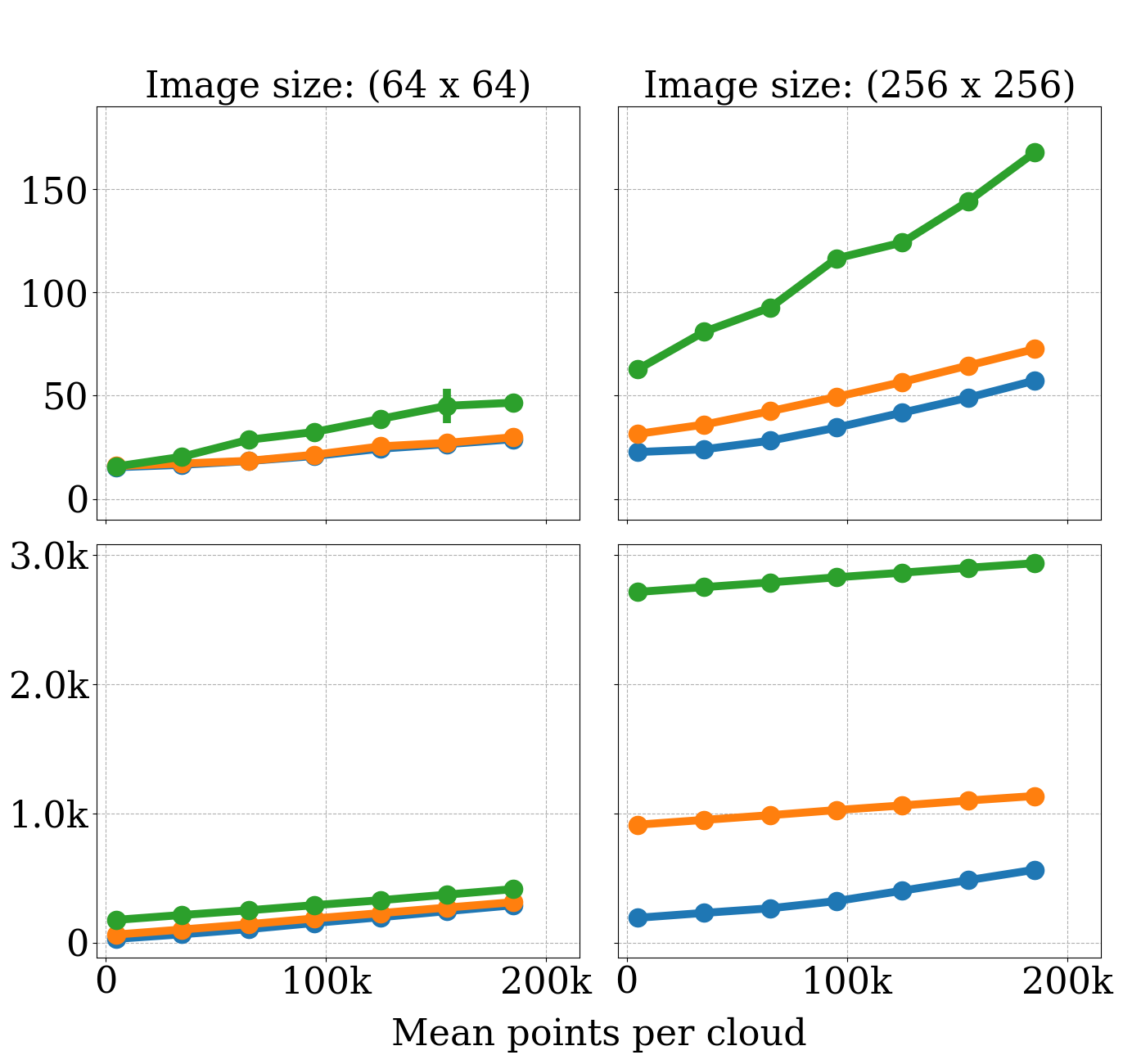}
\caption{\textbf{Alpha} heterogeneous}
\label{fig:alpha_het}
\end{subfigure}
\begin{subfigure}{0.25\textwidth}
\centering
\includegraphics[height=4.9cm]{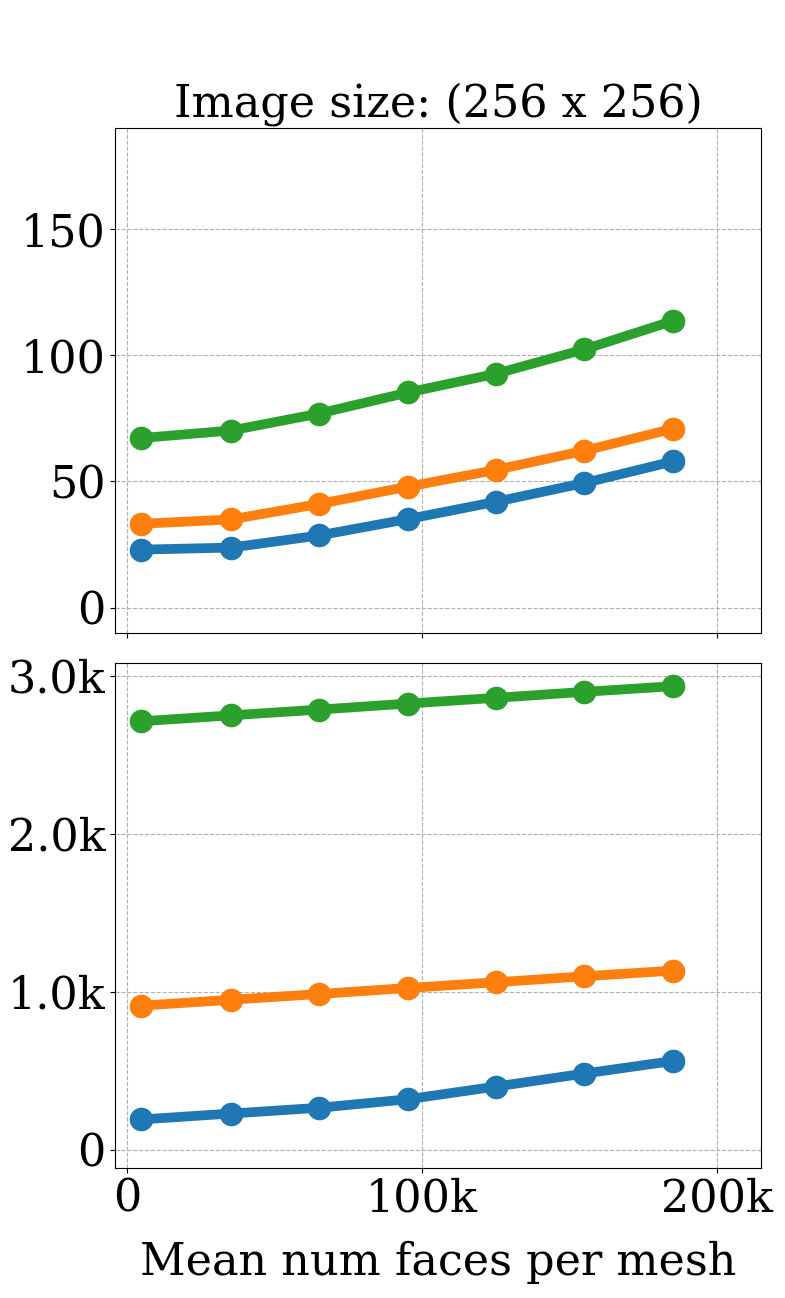}
\caption{\textbf{Norm} heterogeneous}
\label{fig:norm_het}
\end{subfigure}
\vspace{-1mm}
\caption{Benchmarks for PyTorch3D's point cloud render with \emph{Alpha} and \emph{Norm weighted} compositing. We use a batch size of 8, two image sizes (64 \& 256) and three values for the number of faces per pixel K (10, 50, 150). All benchmarks are for forward and backward.}
 \label{fig:benchmark_point_render}
 \vspace{-5mm}
\end{figure}

We benchmark our point cloud renderer by sampling points from the surface of random ShapeNet meshes, then rendering silhouettes using our two blending functions.
We vary the point cloud size, points per pixel ($K=10,50,150$), and image size (64, 256).
Results are shown in Figure~\ref{fig:benchmark_point_render}.

Our renderer is efficient: rendering a batch of 8 point clouds with $200k$ points each to a batch of $256\times256$ images with $K=50$ points per pixel takes about 75ms and uses just over 1GB of GPU memory, making it feasible to use the renderer as a differentiable layer when training neural networks.
Comparing Figures~\ref{fig:alpha_hom} and \ref{fig:alpha_het} shows similar performance when rendering homogenous and heterogeneous batches of comparable size.
Comparing Figures~\ref{fig:alpha_het} and \ref{fig:norm_het} shows that both blending methods have similar memory usage, but \emph{Norm} is up to 25\% faster for large $K$ since it omits the inner cumulative product.
Point cloud rendering is generally more efficient than mesh rendering since it requires fewer computations per primitive during rasterization.

\pagebreak

\section{Experiments}
\label{sec:exp}
\vspace{-1mm}
Extending supervised learning into 3D is challenging due to the difficulty of obtaining 3D annotations. Extracting 3D via weakly or unsupervised approaches can unlock exciting applications, such as novel view synthesis, 3D content creation for AR/VR and more. Differentiable rendering makes 3D inference via 2D supervision possible. 
In this section, we experiment with unsupervised 3D shape prediction using PyTorch3D.
At test time, models predict an object's 3D shape (point cloud or mesh) from a single RGB image.
During training they receive no 3D supervision, instead relying on re-projection losses via differentiable rendering.
We compare to SoftRas~\cite{liu2019soft} and demonstrate superior shape prediction and speed.
The efficiency of our renderer allows us to scale to larger images and more complex meshes, setting a new state-of-the-art for unsupervised 3D shape prediction.

\mypar{Dataset.}
We experiment on ShapeNetCoreV1~\cite{shapenet}, using the rendered images and train/test splits from \cite{choy2016r2n2}.
Images are 137$\times$137, and portray instances from 13 object categories from various viewpoints.
There are roughly 840K train and 210K test images; we reserve $5\%$ of train images for validation.

\mypar{Metrics.}
We follow \cite{wang2018pixel2mesh,gkioxari2019mesh} for evaluating 3D meshes.
We sample 10k points uniformly at random from the surface of predicted and ground-truth meshes.
These are compared using Chamfer distance, normal consistency, and $F_1^\tau$ score for various distance thresholds $\tau$.
Refer to \cite{gkioxari2019mesh} for more details. To fairly extend evaluation to point cloud models, we predict 10k points per cloud.

\vspace{-2mm}
\subsection{Mesh prediction with differentiable rendering}
\vspace{-1mm}
\label{sec:mesh_render}
In this task, we predict 3D object meshes from 2D image inputs with 2D silhouette supervision.
Following \cite{liu2019soft}, we use a 2-view training setup: for each object image on the minibatch we include its corresponding view under a random known transformation. 
At test time, all models take as input a single image and directly predict the 3D object mesh \emph{in camera coordinates}. 
Inspired by recent advances in supervised shape prediction~\cite{wang2018pixel2mesh, gkioxari2019mesh}, we explore the following model designs:

\begin{figure}  
  \begin{subfigure}{0.65\textwidth}
  \centering
  \begin{minipage}{0.1\linewidth} \small Input \end{minipage}
  \begin{minipage}{0.2\linewidth} \centering \small Sphere FC \end{minipage}
  \begin{minipage}{0.2\linewidth} \centering \small Sphere GCN \end{minipage}
  \begin{minipage}{0.2\linewidth} \centering \small High Res Sphere GCN \end{minipage}
  \begin{minipage}{0.2\linewidth} \centering \small Voxel GCN \end{minipage} \\*
  \includegraphics[trim=21 21 21 21,clip,width=0.1\linewidth]{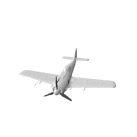}
  \includegraphics[trim=30 30 30 30,clip,width=0.1\linewidth]{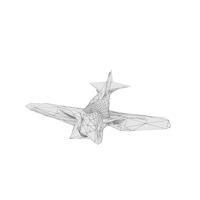}
  \includegraphics[trim=30 30 30 30,clip,width=0.1\linewidth]{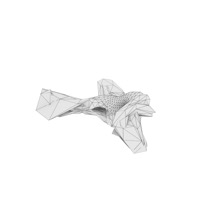}
  \includegraphics[trim=30 30 30 30,clip,width=0.1\linewidth]{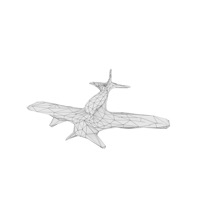}
  \includegraphics[trim=30 30 30 30,clip,width=0.1\linewidth]{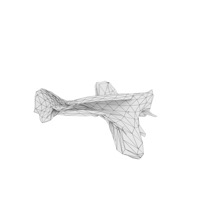}
  \includegraphics[trim=30 30 30 30,clip,width=0.1\linewidth]{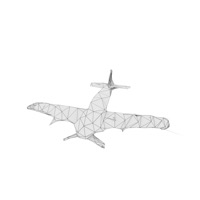}
  \includegraphics[trim=30 30 30 30,clip,width=0.1\linewidth]{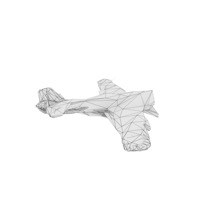}
  \includegraphics[trim=30 30 30 30,clip,width=0.1\linewidth]{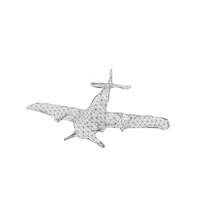}
  \includegraphics[trim=30 30 30 30,clip,width=0.1\linewidth]{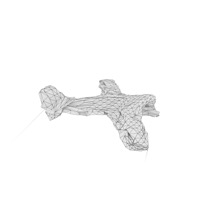} 

  \includegraphics[trim=11 11 11 11,clip,width=0.1\linewidth]{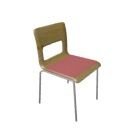}
  \includegraphics[trim=15 15 15 15,clip,width=0.1\linewidth]{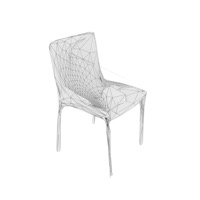}
  \includegraphics[trim=15 15 15 15,clip,width=0.1\linewidth]{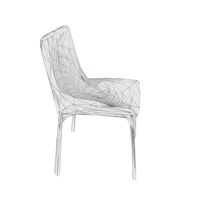}
  \includegraphics[trim=15 15 15 15,clip,width=0.1\linewidth]{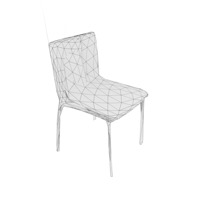}
  \includegraphics[trim=15 15 15 15,clip,width=0.1\linewidth]{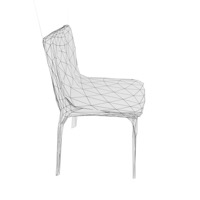}
  \includegraphics[trim=15 15 15 15,clip,width=0.1\linewidth]{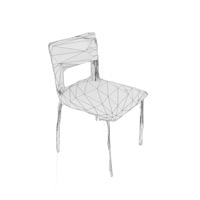}
  \includegraphics[trim=15 15 15 15,clip,width=0.1\linewidth]{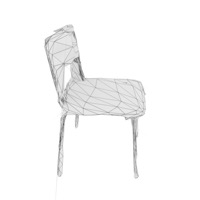}
  \includegraphics[trim=15 15 15 15,clip,width=0.1\linewidth]{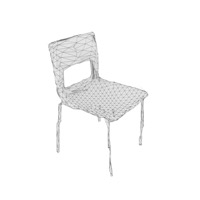}
  \includegraphics[trim=15 15 15 15,clip,width=0.1\linewidth]{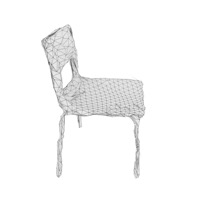} 

  \includegraphics[trim=21 11 21 21,clip,width=0.1\linewidth]{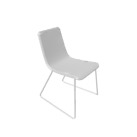}
  \includegraphics[trim=30 15 30 30,clip,width=0.1\linewidth]{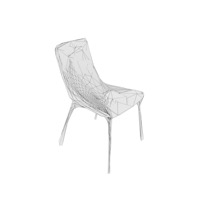}
  \includegraphics[trim=30 15 30 30,clip,width=0.1\linewidth]{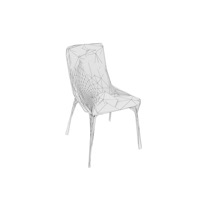}
  \includegraphics[trim=30 15 30 30,clip,width=0.1\linewidth]{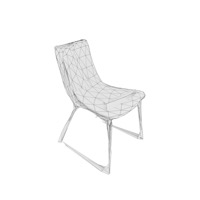}
  \includegraphics[trim=30 15 30 30,clip,width=0.1\linewidth]{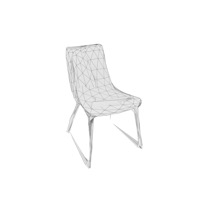}
  \includegraphics[trim=30 15 30 30,clip,width=0.1\linewidth]{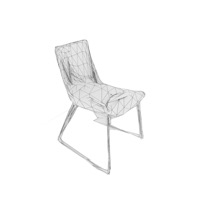}
  \includegraphics[trim=30 15 30 30,clip,width=0.1\linewidth]{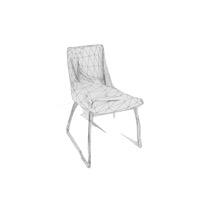}
  \includegraphics[trim=30 15 30 30,clip,width=0.1\linewidth]{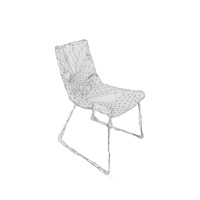}
  \includegraphics[trim=30 15 30 30,clip,width=0.1\linewidth]{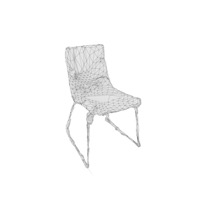} 

  \includegraphics[trim=0 0 0 11,clip,width=0.1\linewidth]{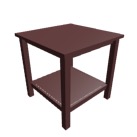}
  \includegraphics[trim=0 0 0 15,clip,width=0.1\linewidth]{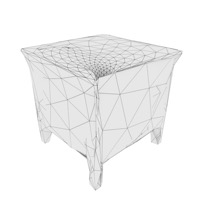}
  \includegraphics[trim=0 0 0 15,clip,width=0.1\linewidth]{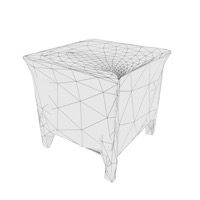}
  \includegraphics[trim=0 0 0 15,clip,width=0.1\linewidth]{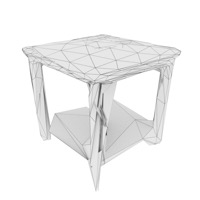}
  \includegraphics[trim=0 0 0 15,clip,width=0.1\linewidth]{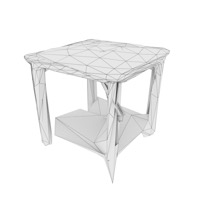}
  \includegraphics[trim=0 0 0 15,clip,width=0.1\linewidth]{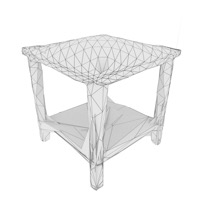}
  \includegraphics[trim=0 0 0 15,clip,width=0.1\linewidth]{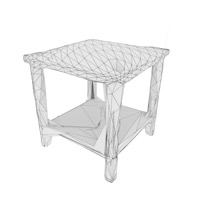}
  \includegraphics[trim=0 0 0 15,clip,width=0.1\linewidth]{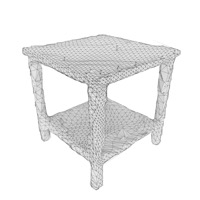}
  \includegraphics[trim=0 0 0 15,clip,width=0.1\linewidth]{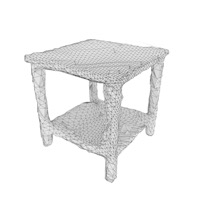} 
  \vspace{-1mm}
  \caption{Silhouette Mesh Rendering}
  \label{fig:silh_mesh_render}
  \end{subfigure}
  \hfill
  \begin{subfigure}{.32\textwidth}
  \centering
  \vspace{1.5mm} 
  \begin{minipage}{0.28\linewidth} \centering \small Input \end{minipage}
  \begin{minipage}{0.36\linewidth}\centering  \small Sphere GCN \end{minipage} 
  \begin{minipage}{0.32\linewidth}\centering  \small Voxel GCN \end{minipage} \\*
  \includegraphics[trim=0 26 0 13,clip,width=0.32\linewidth]{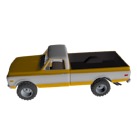}
  \includegraphics[trim=0 38 0 19,clip,width=0.32\linewidth]{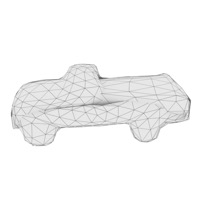}
  \includegraphics[trim=0 38 0 19,clip,width=0.32\linewidth]{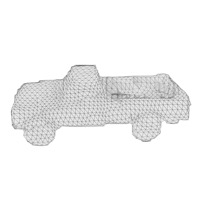} \\
  
 \FramedBox{0.19\linewidth}{0.32\linewidth}{\small Flat}
 \includegraphics[trim=0 26 0 26,clip,width=0.32\linewidth]{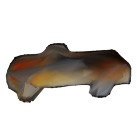}
 \includegraphics[trim=0 26 0 26,clip,width=0.32\linewidth]{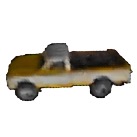} \\
 
 \FramedBox{0.19\linewidth}{0.32\linewidth}{\small Phong}
 \includegraphics[trim=0 26 0 26,clip,width=0.32\linewidth]{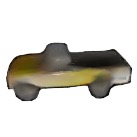}
 \includegraphics[trim=0 26 0 26,clip,width=0.32\linewidth]{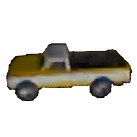} \\
 
 \FramedBox{0.19\linewidth}{0.32\linewidth}{\small Gouraud}
 \includegraphics[trim=0 26 0 26,clip,width=0.32\linewidth]{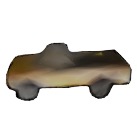}
 \includegraphics[trim=0 26 0 26,clip,width=0.32\linewidth]{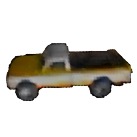}
 \caption{Textured Mesh Rendering}
 \label{fig:text_mesh_render}
  \end{subfigure}
  \vspace{-2mm}
  \caption{Mesh predictions via (a) \textbf{silhouette} and (b) \textbf{textured rendering} on ShapeNet test. For (a), we show the input image, the prediction and an additional view for each model. For (b), we show the input image, the predicted shapes and the predicted textures for each shader and model - the predicted shapes across shaders are very similar, so we show one per model.}
  \label{fig:silh_text_mesh_render}
  \vspace{-5mm}
\end{figure}

\mypar{Sphere FC.} Closely following \cite{liu2019soft}, this model learns to deform an initial sphere template with 642 vertices and 1280 faces. The input image is encoded via a CNN backbone followed by two fully connected layers, each of 1024 dimensions, which cast offset predictions for each vertex.

\mypar{Sphere GCN.} Inspired by recent advances in~\cite{wang2018pixel2mesh, gkioxari2019mesh}, this model uses graph convolutions. Each vertex pools its features from the output of the backbone indexed by its 2D projection. 
A set of graph convs on the mesh graph predict vertex offsets. Similar to Sphere FC, this model deforms a sphere mesh. Unlike Sphere FC, image to feature alignment is preserved. PyTorch3D's scale efficiency allows us to train a High Res variant which deforms an even larger sphere (2562 verts, 5120 faces).

\mypar{Voxel GCN.} The above models predict shapes homeomorphic to spheres, and are not able to capture varying shape topologies. Recently, Mesh R-CNN~\cite{gkioxari2019mesh} shows that coarse voxel predictions capture instance-specific object topologies. These low fidelity voxel predictions perform poorly as they can't reconstruct fine structures or smooth surfaces, but when refined with 3D mesh supervision, they become state-of-the-art. To demonstrate PyTorch3D's flexibility with varying shape topologies and heterogeneous batches, we train a model which makes voxel predictions and refines them via a sequence of graph convs.  At train time, this model uses coarse $48^3$ voxel supervision similar to \cite{gkioxari2019mesh}, but unlike \cite{gkioxari2019mesh} it does not use 3D mesh supervision.

All models minimize the objective $\mathcal{L} = \mathcal{L}_s + \lambda_l \mathcal{L}_l + \lambda_e \mathcal{L}_e$. $\mathcal{L}_s$ is the negative intersection over union between rendered and ground truth 2D silhouette, as in \cite{liu2019soft}. $\mathcal{L}_l$ and $\mathcal{L}_e$ are Laplacian and edge length mesh regularizers, respectively, ensuring that meshes are smooth. We set $\lambda_l$=19 and $\lambda_e$=0.2.

All model variants use a ResNet50~\cite{he2016deep} backbone, initialized with ImageNet weights. We follow the training schedule from~\cite{gkioxari2019mesh}; we use Adam with a constant learning rate of $10^{-4}$ for 25 epochs. We use a 64 batch size across 8 V100 GPUs (8 images per GPU). Sphere GCN and Voxel GCN use a sequence of 3 graph convs, each with 512 dimensions. The voxel head in Voxel GCN predicts $48^3$ voxels via a 4 layer CNN, identical to \cite{gkioxari2019mesh}. For all models, the inputs are 137$\times$137 images from \cite{choy2016r2n2}. We report performance using our mesh renderer and compare to SoftRas~\cite{liu2019soft}.

Table~\ref{tab:silh_mesh} shows our results. We compare to the supervised state-of-the-art Mesh R-CNN~\cite{gkioxari2019mesh} and its Voxel Only variant, the latter being a direct comparison to Voxel GCN as both use the same 3D supervision -- coarse voxels but no meshes. 
We render at 64$\times$64, similar to SoftRas~\cite{liu2019soft}, and push to even higher rendering resolution at 128$\times$128.
Figure~\ref{fig:silh_mesh_render} compares the models qualitatively. 

\newcommand{\meanstd}[2]{#1\scalebox{0.65}{$\pm#2$}}
\begin{table}
  \small
  \centering
  \scalebox{0.95}{
  \begin{tabular}{C{.11\textwidth}C{.02\textwidth}C{.03\textwidth}C{.03\textwidth}C{.08\textwidth}C{.07\textwidth}C{.05\textwidth}C{.03\textwidth}C{.03\textwidth}C{.03\textwidth}C{.07\textwidth}C{.05\textwidth}}
  	\toprule
	\multicolumn{1}{c}{Model}   & \multicolumn{2}{c}{3D Superv.}    & \multicolumn{2}{c}{Renderer} &  \multicolumn{5}{c}{Metrics} & \multicolumn{2}{c}{Mesh Size}  \\
	 \cmidrule(r){1-1} \cmidrule(r){2-3}  \cmidrule(r){4-5} \cmidrule(r){6-10} \cmidrule(r){11-12} 

		\multicolumn{1}{C{.12\textwidth}}{Net}         	& Vox & Mesh  & Size & Engine   & Ch. ($\downarrow$) & Nrml  & $F_1^{0.1}$ & $F_1^{0.3}$ & $F_1^{0.5}$  & $|V|$ & $|F|$ \\
		\midrule
		\midrule
		\multicolumn{1}{L{.12\textwidth}}{Sphere FC}    & \bul  & \bul  & 64         & SoftRas             & 1.475    & 0.691    & 25.5    & 68.4    & 82.3   & 642 & 1280  \\
             			        				    			& \bul  & \bul  &              & PyTorch3D  	 & 0.989    & 0.696    & 26.4    & 69.9    & 83.5   & 642 & 1280 \\
             			        				    			& \bul  & \bul  &128       & SoftRas  	     	 & 0.346    & 0.700    & 26.1    & 70.6    & 85.2   & 642 & 1280 \\
             							    			& \bul  & \bul  &             & PyTorch3D 	 & 0.313    & 0.699    & \textcolor{blue}{\bf 27.6}    & 72.5    & 86.6   & 642 & 1280 \\
		\midrule
		\multicolumn{1}{L{.12\textwidth}}{Sphere GCN} 	& \bul  & \bul  & 64          & SoftRas   	& 0.316   & \textcolor{blue}{\bf 0.713}  & 24.4   & 70.2   & 85.8     & 642 & 1280 \\
             			        				    			& \bul  &  \bul  &              & PyTorch3D 	& 0.296   & 0.703     & 24.8   & 71.3   & 86.5     & 642 & 1280 \\
             							    			& \bul  &  \bul  & 128       & SoftRas   	&  0.301  & 0.709     & 26.1   & 71.9   & 86.5     & 642 & 1280 \\
             							    			& \bul  & \bul   &              & PyTorch3D 	& 0.293   & 0.709   & 26.6   & 72.6   & 86.9     & 642 & 1280 \\
		\midrule
		\multicolumn{1}{L{.12\textwidth}}{High Res Sphere GCN}   & \multirow{2}{*}{\bul} & \multirow{2}{*}{\bul}   & \multirow{2}{*}{128}  & \multirow{2}{*}{PyTorch3D} & \multirow{2}{*}{\textcolor{blue}{\bf 0.281}} & \multirow{2}{*}{0.696} & \multirow{2}{*}{26.7}  &  \multirow{2}{*}{\textcolor{blue}{\bf 73.8}} & \multirow{2}{*}{\textcolor{blue}{\bf 87.8}} & \multirow{2}{*}{2562} &  \multirow{2}{*}{5120}   \\
	        \midrule
		\multicolumn{1}{L{.12\textwidth}}{Voxel GCN}    & \checkmark   & \bul  & 64        & SoftRas   		& 0.293  & 0.656     & 24.5    & 71.1 	& 87.2  & \meanstd{1947}{923} & \meanstd{3895}{1851}    \\
             			         			    			& \checkmark   & \bul  &             & PyTorch3D 		& \textcolor{purple}{\bf 0.267}  & 0.675     & 26.1    & 73.3 	& \textcolor{purple}{\bf 88.5}  & \meanstd{1932}{935} & \meanstd{3866}{1873}   \\
             							    			& \checkmark   & \bul  & 128   	 & SoftRas   		& 0.276  & 0.675     & \textcolor{purple}{\bf 26.2}    & 72.6 	& 87.9  & \meanstd{1918}{928} & \meanstd{3837}{1860}   \\
             							    			& \checkmark   & \bul  &             & PyTorch3D 		& 0.277  & \textcolor{purple}{\bf 0.687}      & \textcolor{purple}{\bf 26.2}   & \textcolor{purple}{\bf 73.4}  & 87.8   & \meanstd{1951}{949} & \meanstd{3903}{1901}    \\
	         \midrule
		\multicolumn{1}{L{.15\textwidth}}{Voxel Only~\cite{gkioxari2019mesh}}  	& \checkmark  & \bul    	    & n/a  & n/a    & 0.916   & 0.595    & 7.70    & 33.1    & 54.9  & \meanstd{2433}{925} & \meanstd{4877}{1856}  \\   
		\multicolumn{1}{L{.17\textwidth}}{Mesh R-CNN~\cite{gkioxari2019mesh}} & \checkmark   & \checkmark & n/a  & n/a    & 0.171    & 0.713   & 35.1    & 82.6    & 93.2  & \meanstd{2292}{902} & \meanstd{4598}{1812}  \\  
   \bottomrule
  \end{tabular}
  } 
  \vspace{0.5mm}
   \caption{Mesh reconstruction via \textbf{silhouette rendering} on ShapeNet test with PyTorch3D and SoftRas~\cite{liu2019soft}. We compare to state-of-the-art Mesh R-CNN~\cite{gkioxari2019mesh}, trained with voxel \& mesh supervision, and its Voxel Only variant, trained with voxel supervision. We highlight best metrics for models trained without any 3D supervision (\textcolor{blue}{\bf blue}) and with voxel supervision (\textcolor{purple}{\bf red}).}
  \label{tab:silh_mesh}
  \vspace{-7mm}
\end{table}
\begin{table}
  \small
  \centering
  \scalebox{0.9}{
  \begin{tabular}{cccccccccc}
  	\toprule
	\multicolumn{1}{c}{Model}     & \multicolumn{2}{c}{Renderer}   & \multicolumn{7}{c}{Metrics}   \\
	 \cmidrule(r){1-1} \cmidrule(r){2-3}  \cmidrule(r){4-10}

		Net          		& Size &  Shading   & Chamfer ($\downarrow$) &Normal & $F_1^{0.1}$ & $F_1^{0.3}$ & $F_1^{0.5}$ & $L_1^{\textrm{\tiny fg}}$ ($\downarrow$) & $L_1^{\textrm{\tiny bg}}$ ($\downarrow$)  \\
		\midrule
		\midrule
		\multicolumn{1}{l}{Sphere GCN}  & 64 	  &  Flat   		& 0.315  &   0.689  &  24.1   &  \textcolor{blue}{\bf 70.9} &  86.3 &  0.0217 & 0.0031    \\
             			         			  & 		  &  Phong		& 0.309  &    \textcolor{blue}{\bf 0.703}  &  24.1   & 70.2 &  85.9 &   \textcolor{blue}{\bf 0.0173} &  \textcolor{blue}{\bf 0.0023} \\
             							  &  		  &  Gouraud  	&  \textcolor{blue}{\bf 0.302} &    0.702  &   \textcolor{blue}{\bf 24.4}   &  \textcolor{blue}{\bf 70.9}  &  \textcolor{blue}{\bf 86.6} & 0.0180  & 0.0024   \\
		\midrule
		\multicolumn{1}{l}{Voxel GCN}    & 64 	&  Flat   		& \textcolor{purple}{\bf 0.270}  & 0.678    & \textcolor{purple}{\bf 26.4}    & \textcolor{purple}{\bf 73.0} & \textcolor{purple}{\bf 88.2}   &  0.0128 & \textcolor{purple}{\bf 0.0020}    \\
             			         			  & 		 &  Phong		& 0.272  & \textcolor{purple}{\bf 0.694}     & 25.7    & 72.4 & 87.9  & \textcolor{purple}{\bf 0.0126} & 0.0021  \\
             							  &  		 & Gouraud  	& 0.302  & 0.687     & 24.3    & 69.8 & 86.2  & 0.0128 & 0.0021   \\
   \bottomrule
  \end{tabular}
  } 
  \vspace{0.5mm}
  \caption{Mesh and texture reconstruction via \textbf{textured rendering} on ShapeNet test. $L_1^{\textrm{\tiny fg}}$ \& $L_1^{\textrm{\tiny bg}}$ measure the reconstruction accuracy of the foreground object and background, respectively.}
  \label{tab:text_mesh}
  \vspace{-7mm}
\end{table}

From Table~\ref{tab:silh_mesh} we observe:
(a) Compared to SoftRas, PyTorch3D achieves on par or better performance across models. This validates the design of the Pytorch3D renderer and proves that rendering the K closest faces, instead of all, does not hurt performance,
(b) Even though Sphere FC \& Sphere GCN deform the same sphere, Sphere GCN is superior across renderers. Figure~\ref{fig:silh_mesh_render} backs this claim qualitatively. Unlike Sphere GCN, Sphere FC is sensitive to rendering size (row 4 vs 6),
(c) High Res Sphere GCN significantly outperforms Sphere GCN both quantitatively (row 10 vs 11) and qualitatively (Figure~\ref{fig:silh_mesh_render}) showing the advantages of PyTorch3D's scale efficiency,
(d) Voxel GCN significantly outperforms Voxel Only, both trained with the same 3D supervision. As mentioned in~\cite{gkioxari2019mesh}, Voxel Only performs poorly, since voxel predictions are coarse and fail to capture fine shapes. Voxel GCN improves all metrics and reconstructs fine structures with complex topologies as shown in Figure~\ref{fig:silh_mesh_render}. Here, PyTorch3D's efficiency results in a 2$\times$ training speedup compared to SoftRas.

\vspace{-1mm}
\begin{wraptable}{r}{6.45cm}
\vspace{-5mm}
  \small
  \centering
  \vspace{2mm}
  \scalebox{0.99}{
  \begin{tabular}{C{.03\textwidth}C{.02\textwidth}C{.05\textwidth}C{.04\textwidth}C{.03\textwidth}C{.03\textwidth}C{.03\textwidth}}
  	\toprule

		 Net         					    & K  & Ch.($\downarrow$)  & Nrml & $F_1^{0.1}$ & $F_1^{0.3}$ & $F_1^{0.5}$  \\
		\midrule
		\midrule
		\multicolumn{1}{l}{Sphere}   	   & 20   		& 0.294  	 & 0.697     & \textcolor{blue}{\bf 27.2}     & 72.2 & 86.8     \\
             	\multicolumn{1}{l}{GCN}		   & 50		& \textcolor{blue}{\bf 0.293}  	 & 0.709     & 26.6     & \textcolor{blue}{\bf 72.6} & \textcolor{blue}{\bf 86.9}     \\
             							    & 100  		& 0.294 	 & 0.708     & 26.9     & 72.3 & \textcolor{blue}{\bf 86.9}    \\
							            & 150             &  0.314     & \textcolor{blue}{\bf 0.716}     & 25.2     & 71.1  & 86.1    \\
		\multicolumn{1}{l}{Voxel}     	   & 20   		& 0.317  	& 0.642      & 24.1     & 68.4  & 85.3      \\
		 \multicolumn{1}{l}{GCN}               & 50		& \textcolor{purple}{\bf 0.277}  	& \textcolor{purple}{\bf 0.687}      & \textcolor{purple}{\bf 26.2}     & \textcolor{purple}{\bf 73.4}  & \textcolor{purple}{\bf 87.8}   \\
             							    & 100  		& 0.282  	& 0.669      & 25.5     & 71.6  & 87.3   \\
							            & 150		& 0.285    & 0.674      & 25.9     & 72.2  & 87.2   \\      
   \bottomrule
  \end{tabular}
  } 
   \caption{Varying K for mesh reconstruction.}
  \label{tab:silh_mesh_K}
  \vspace{-2mm}
\end{wraptable}
\mypar{Varying K}
As described in Section~\ref{sec:bench}, the PyTorch3D renderer exposes the K nearest faces per pixel.
We experiment with different values of K for Sphere GCN \& Voxel GCN at a 128$\times$128 resolution in Table~\ref{tab:silh_mesh_K}.
We observe that K=50 results in best performance for both models (Chamfer 0.293 for Sphere GCN \& 0.277 for Voxel GCN).
More interestingly, a small value for K(=20) works well for smaller meshes (Sphere GCN) but results in a performance drop for larger meshes (Voxel GCN) with the same output image size. This is expected since for the same K, for a larger mesh, fewer faces per pixel are rendered in proportion to the mesh size. Finally, increasing K does not improve models further. This empirically validates our design of rendering a fixed finite number of faces per pixel.

\mypar{Textured rendering}
In addition to shapes, we reconstruct object textures by extending the above models to predict per vertex $(r,g,b)$ values using textured rendering. The models are trained with an additional $L_1$ loss between the rendered and the ground truth image. Table~\ref{tab:text_mesh} shows our analysis. Simultaneous shape and texture prediction is harder, yet our models achieve high reconstruction quality for both shapes (compared to Table~\ref{tab:silh_mesh}) and textures. Figure~\ref{fig:text_mesh_render} shows qualitative results.

\vspace{-2mm}
\subsection{Point cloud prediction with differentiable rendering}
\vspace{-2mm}
To show the effectiveness of the point cloud renderer, we train unsupervised point cloud models.
Our model, called Point Align, deforms 10k points sampled from a sphere by pooling backbone features and predicting per point offsets.
Table~\ref{tab:silh_cloud} shows our analysis. Point Align slightly improves shape metrics compared to meshes (Tables~\ref{tab:silh_cloud} vs~\ref{tab:silh_mesh}). As with meshes, a finite K(=100) performs best while increasing K does not improve models further. We reconstruct texture by predicting additional $(r,g,b)$ values per point via textured rendering (Table~\ref{tab:text_cloud}). Figure~\ref{fig:silh_text_cloud_render} shows qualitative results. Finally, we compare Point Align to the supervised PSG~\cite{fan2017point} baseline in Table~\ref{tab:cloud_vs_psg}; we significantly improve $F_1$ and show slightly worse Chamfer, which is expected as PSG directly minimizes Chamfer with 3D supervision, while our model is not trained to minimize any 3D metric and uses no supervision. 

 \begin{table}
 \begin{minipage}[t]{0.43\linewidth}
  \small
  \centering
  \scalebox{1.0}{
  \begin{tabular}{C{.05\textwidth}C{.06\textwidth}C{.17\textwidth}C{.06\textwidth}C{.06\textwidth}C{.06\textwidth}}
  	\toprule
	\multicolumn{2}{c}{Renderer}   & \multicolumn{4}{c}{Metrics}   \\
	 \cmidrule(r){1-2}  \cmidrule(r){3-6}

		 Size & K  & Ch. ($\downarrow$)  & $F_1^{0.1}$ & $F_1^{0.3}$ & $F_1^{0.5}$\\
		\midrule
		\midrule
		64		& 20   		& 0.738  	 & 12.8       & 50.6     & 68.0     \\
             		        & 50			& 0.451  	 & 18.1       & 62.9     & 79.0     \\
             			& 100  		& 0.289 	 & 23.4       & 73.3     & 87.1     \\
				& 150      		&  0.289     & 23.9       & 73.5     & 87.3      \\
		128           & 20   	        & 0.623  	& 13.7        & 53.1     & 71.2      \\
             		        & 50			& 0.398  	& 19.2        & 65.6     & 81.3      \\
             			& 100  		& \textcolor{blue}{\bf 0.272} 	& \textcolor{blue}{\bf 25.3}        & \textcolor{blue}{\bf 75.2}     & \textcolor{blue}{\bf 88.0}     \\
				& 150		& 0.280    & 25.1        & 74.6     & 87.8     \\      
   \bottomrule
  \end{tabular}
  } 
  \vspace{1.9mm}
  \caption{Point cloud reconstruction via \textbf{silhouette rendering} on ShapeNet test for different rendering resolutions and K.}
  \label{tab:silh_cloud}
  \vspace{-4mm}
 \end{minipage} 
 \hspace{3mm}
 \begin{minipage}[t]{0.55\linewidth}
  \begin{minipage}{0.6\linewidth}
  \begin{minipage}{0.3\linewidth} \centering \small Input \end{minipage}
  \begin{minipage}{0.6\linewidth} \centering \small Point Align \end{minipage} \\*
  \includegraphics[trim=21 21 21 11,clip,width=0.3\linewidth]{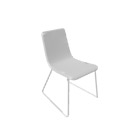}
  \includegraphics[trim=30 30 30 15,clip,width=0.3\linewidth]{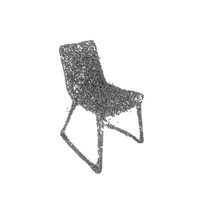}
  \includegraphics[trim=30 30 30 15,clip,width=0.3\linewidth]{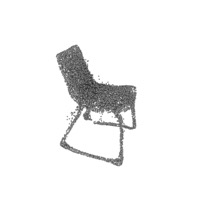} \\
  \includegraphics[trim=21 21 21 21,clip,width=0.3\linewidth]{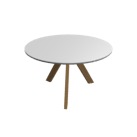}
  \includegraphics[trim=30 30 30 30,clip,width=0.3\linewidth]{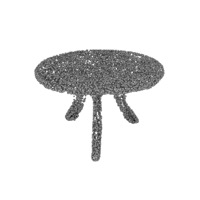} 
  \includegraphics[trim=30 30 30 30,clip,width=0.3\linewidth]{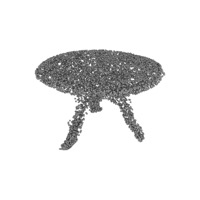}
 \end{minipage}
 \hfill
 \begin{minipage}{0.37\linewidth}
  \FramedBox{0.4\linewidth}{0.2\linewidth}{\small Input}
  \includegraphics[trim=0 20 0 31,clip,width=0.73\linewidth]{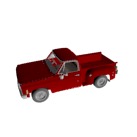} 
  \FramedBox{0.4\linewidth}{0.2\linewidth}{\small Alpha}
  \includegraphics[trim=0 20 0 31,clip,width=0.73\linewidth]{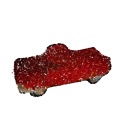}  
  \FramedBox{0.4\linewidth}{0.22\linewidth}{\small Norm}
  \includegraphics[trim=0 20 0 31,clip,width=0.73\linewidth]{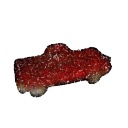} 
 \end{minipage}
 \begin{minipage}{0.6\linewidth} \centering \small (a) \end{minipage}
 \begin{minipage}{0.35\linewidth} \centering \small (b) \end{minipage} 
 \vspace{0.5mm}
 \captionof{figure}{Point cloud predictions via (a) \textbf{silhouette} and (b) \textbf{textured rendering} on ShapeNet test.}
 \label{fig:silh_text_cloud_render}
 \end{minipage}
 \vspace{-6mm}
 \end{table}

\begin{table}
  \small
  \begin{minipage}[b]{.6\linewidth}
   \scalebox{0.9}{
  \begin{tabular}{C{.06\textwidth}C{.06\textwidth}C{.08\textwidth}C{.08\textwidth}C{.05\textwidth}C{.05\textwidth}C{.05\textwidth}C{.1\textwidth}C{.1\textwidth}}
  	\toprule
	\multicolumn{3}{c}{Renderer}        & \multicolumn{6}{c}{Metrics}   \\
	\cmidrule(r){1-3} \cmidrule(r){4-9}

		Size & K & Blend &  Ch.($\downarrow$)  & $F_1^{0.1}$ & $F_1^{0.3}$ & $F_1^{0.5}$ & $L_1^{\textrm{\tiny fg}}$($\downarrow$) & $L_1^{\textrm{\tiny bg}}$($\downarrow$)  \\
		\midrule
		\midrule
		128  		& 100   & Alpha      	& 0.275  & 25.2       & 74.7     & 87.8  & \textcolor{blue}{\bf 0.0178}  & 0.0051  \\
             			&   	    & Norm  	& \textcolor{blue}{\bf 0.268}  & \textcolor{blue}{\bf 25.7}       & \textcolor{blue}{\bf 75.4}     & \textcolor{blue}{\bf 88.1}  & 0.0187  & \textcolor{blue}{\bf 0.0027}  \\
								    
   \bottomrule
  \end{tabular}
  } 
  \vspace{0.5mm}
   \caption{Point cloud and texture reconstruction via \textbf{textured rendering} on ShapeNet test with two blending functions.}
  \label{tab:text_cloud}
  \end{minipage}
  \hspace{0.6mm}
  \begin{minipage}[b]{.38\linewidth}
  \small
  \centering
   \scalebox{0.99}{
  \begin{tabular}{C{.08\textwidth}C{.03\textwidth}C{.11\textwidth}C{.06\textwidth}C{.06\textwidth}}
  	\toprule
		\multicolumn{2}{c}{Net \hspace{5mm}  3D superv} &  Ch.($\downarrow$)  & $F_1^{\tau}$ & $F_1^{2\tau}$ \\
		\midrule
		\midrule
		\multicolumn{1}{l}{PSG~\cite{fan2017point}}  & \checkmark & {\bf 0.593} & 48.6 & 69.8 \\
		\multicolumn{1}{l}{Point Align}    		    & \bul 	  & 0.647  & {\bf 61.0}  & {\bf 74.6} \\							    
   \bottomrule
  \end{tabular}
  } 
  \vspace{0.5mm}
   \caption{Comparison of our unsupervised point cloud model, Point Align, to PSG~\cite{fan2017point} under the non scale-normalized metric (Table 1 in~\cite{gkioxari2019mesh}).}
  \label{tab:cloud_vs_psg}
  \end{minipage}
  \vspace{-6mm}
\end{table}

\section{Broader Impact}
In this paper we introduce PyTorch3D, a library for 3D research which is \emph{fully differentiable} to enable easy inclusion into existing deep learning pipelines, \emph{modular} for fast experimentation and extension, \emph{optimized} and \emph{efficient} to allow scaling to large 3D data sizes, and with \emph{heterogeneous batching} capabilities to support the variable shape topologies encountered in real world use cases. 

Similar to PyTorch or TensorFlow, which provide fundamental tools for deep learning, PyTorch3D is a library of building blocks optimized for 3D deep learning. Frameworks like PyTorch and PyTorch3D provide a platform for solving a plethora of AI problems including semantic or synthesis tasks. Data-driven solutions to such problems necessitate the community's caution regarding their potential impact, especially when these models are being deployed in the real world. In this work, our goal is to provide students, researchers and engineers with the best possible tools to accelerate research at the intersection of 3D and deep learning. To this end, we are committed to support and develop PyTorch3D in adherence with the needs of the academic, research and engineering community.

\section*{Appendix}
\appendix

\section{Sampling Batches for Benchmarks}
\label{app:batch_bench}

The batches of 3D data used in the benchmarks in Section~\ref{sec:bench} were sampled from ShapeNetCoreV1 using a uniform sampling strategy. For meshes, homogeneous batches ($\sigma = 0$; $\sigma$ here denotes the variance in size for elements within the batch) consist of one mesh with the specified number of faces, repeated $B$ times with $B$ being the batch size. For heterogeneous batches ($\sigma > 0$), we sample $B$ values from a uniform distribution with the specified $\mu$ and $\sigma$ in the number of faces per mesh. For each value, we find the mesh in the dataset with number of faces closest to the desired value. 

For point cloud operators, we first sample a random mesh from the dataset and then uniformly sample points from the surface of the mesh. For homogeneous batches, the same number of points is sampled $B$ times. For heterogeneous batches, $B$ values for the number of points in each point cloud is sampled from a uniform distribution with the specified $\mu$ and $\sigma$.

\section{Experiments: Unsupervised shape prediction}
\label{app:exp}

In Section~\ref{sec:exp}, we experiment with unsupervised 3D shape prediction from a single image using the PyTorch3D renderers. Predicting a 3D shape from an input image is ambiguous as infinite 3D shapes can explain a 2D image. In the absence of ground truth supervision, to resolve shape ambiguity we assume 2 views of the object at train time. This setup is also assumed in SoftRas~\cite{liu2019soft}. 

\begin{figure}
  \centering
    \includegraphics[width=0.99\linewidth]{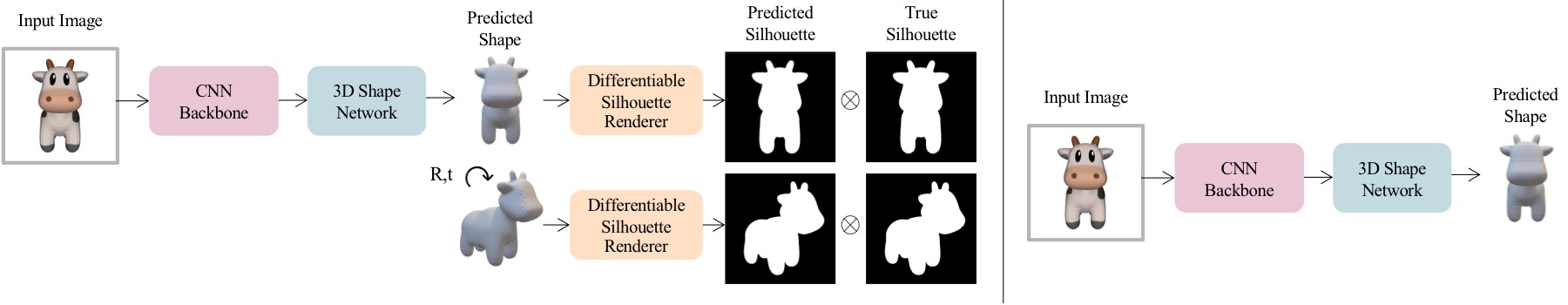}
    \begin{minipage}{0.6\linewidth} \centering \small (a) Train time \end{minipage}
    \begin{minipage}{0.35\linewidth} \centering \small (b) Inference \end{minipage} 
  \caption{System overview for unsupervised shape prediction. (a) Shows the 2-view training setup. During training, an additional view of known rotation $R$ and translation $t$ of the input view is assumed. The predicted shape is transformed by $(R,t)$ and its rendered silhouette is compared against the ground truth one. (b) Shows the system at inference time. The input is a single RGB image and the output is the predicted shape \emph{in camera coordinates}.}
  \label{fig:overview}
\end{figure}

The 2-view training setup is shown in Figure~\ref{fig:overview}(a). When constructing minibatches, for every image input sampled, we assume an additional view under known rotation $R$ and translation $t$. The predicted shape from the input is then transformed by $(R,t)$ and its rendered output is compared against the ground truth silhouette. At test time, the model takes as input a single image and predicts the object's 3D shape \emph{in camera coordinates}, as shown in Figure~\ref{fig:overview}(b). In the case of textured rendering, the textured views are additionally compared at train time. We do the same for point clouds. 

\subsection{Unsupervised mesh prediction}

\begin{figure}
  \begin{subfigure}{1.0\textwidth}
   	\centering
   	\includegraphics[width=0.99\linewidth]{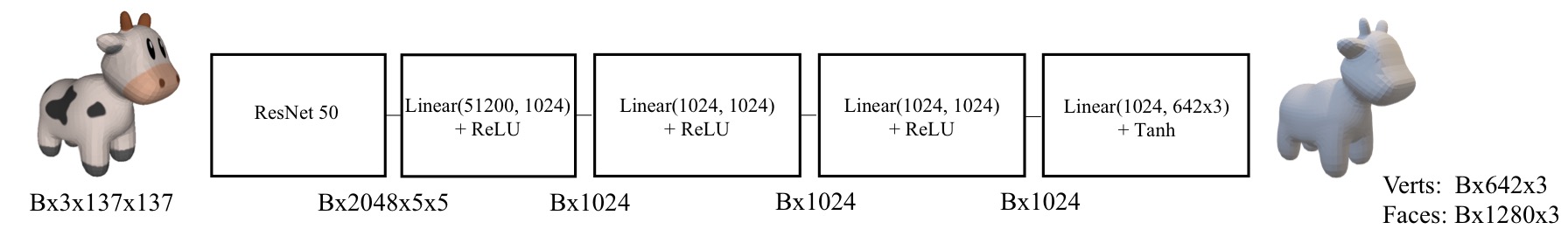}
	\caption{Sphere FC}
	\label{fig:sphere_fc}
  \end{subfigure}
   \begin{subfigure}{1.0\textwidth}
   	\centering
   	\includegraphics[width=0.99\linewidth]{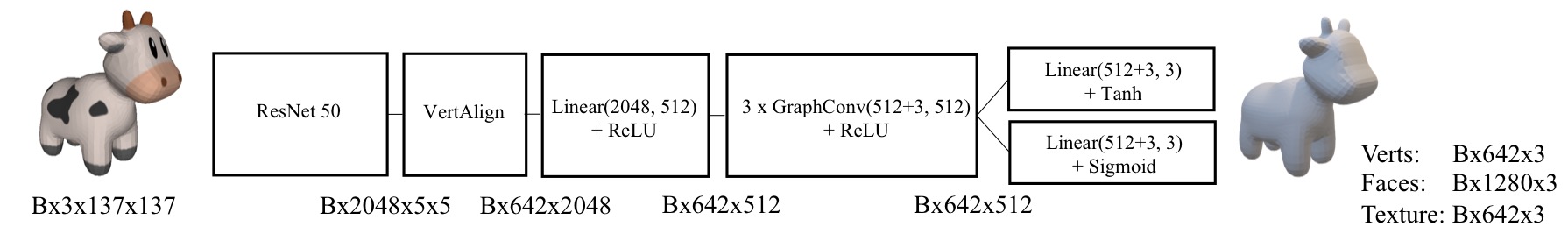}
	\caption{Sphere GCN}
	\label{fig:sphere_gcn}
  \end{subfigure}
   \begin{subfigure}{1.0\textwidth}
   	\vspace{3mm}
   	\centering
   	\includegraphics[width=0.99\linewidth]{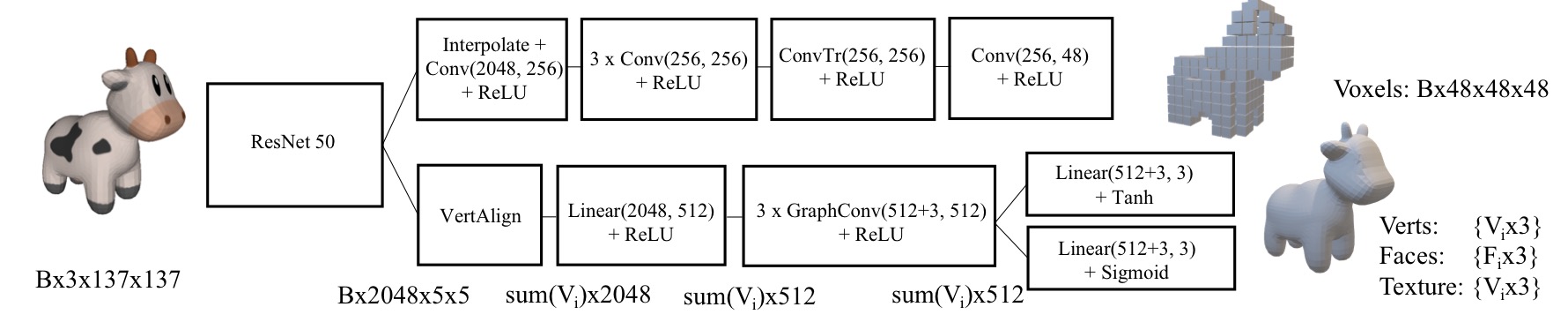}
	\caption{Voxel GCN}
	\label{fig:voxel_gcn}
  \end{subfigure}
  \caption{3D mesh network architectures used in our experiments.}
  \label{fig:mesh_networks}
\end{figure}

In the case of meshes, we experiment with three architectures for the 3D shape network, namely \emph{Sphere FC}, \emph{Sphere GCN} and \emph{Voxel GCN}. The first two deform an initial sphere template and use no 3D shape supervision. The third deforms shape topologies cast by the voxel head and tests the effectiveness of differentiable rendering for varying shapes and connectivities as predicted by a neural network. These topologies are different and more diverse than human-defined or simple genus 0 shapes allowing us to stretch test our renderer. Voxel GCN uses coarse voxel supervision to train the voxel head which learns to predict instance-specific, yet coarse, shapes. Note that Voxel GCN is identical to Mesh R-CNN~\cite{gkioxari2019mesh}, but unlike Mesh R-CNN it uses no mesh supervision. The details of the architectures for all three models are shown in Figure~\ref{fig:mesh_networks}. For each model, we show the network modules as well as the shapes of the intermediate batched tensor outputs. Note that for Voxel GCN, the shapes of the meshes vary within the batch due to the heterogeneity from the voxel predictions. 

\mypar{Losses}
As stated in Section~\ref{sec:exp}, all models optimize the same objective $\mathcal{L} = \mathcal{L}_s + \lambda_l \mathcal{L}_l + \lambda_e \mathcal{L}_e$. 
$\mathcal{L}_s$ is the negative intersection over union between rendered and ground truth 2D silhouette from \cite{liu2019soft}
\begin{equation}
\mathcal{L}_s = 1 - \frac{S_\textrm{pred} \cdot S_\textrm{gt}}{S_\textrm{pred} + S_\textrm{gt} - S_\textrm{pred} \cdot S_\textrm{gt}}
\end{equation}
where $S_\textrm{pred}$ and $S_\textrm{gt}$ are the predicted and ground truth silhouettes. To enforce smoothness in the predicted shapes we also use shape regularizers. We use an edge length regularizer $\lambda_e$ that minimizes the length of the edges in the predicted mesh, identical to Mesh R-CNN~\cite{gkioxari2019mesh}. We also use a laplacian regularizer $\mathcal{L}_l$, defined as follows 
\begin{equation}
\mathcal{L}_l = || L \cdot v ||_1
\end{equation}
where $L$ is the Laplacian matrix of shape $V \times V$ and $v$ are the vertices of shape $V \times 3$. We use PyTorch3D's implementation of this loss that handles heterogeneous batches efficiently. At train time, the loss is averaged across vertices and elements in the batch. This is unlike SoftRas~\cite{liu2019soft}, where it is summed across vertices and across elements in the batch.

\mypar{More discussion on Table~\ref{tab:silh_mesh} and Figure~\ref{fig:silh_mesh_render}}
Table~\ref{tab:silh_mesh} compares performance for different shape networks with PyTorch3D and SoftRas and under two rendering resolutions. We replicate the original experiments in SoftRas~\cite{liu2019soft} which use a Sphere FC model to deform a sphere of 642 vertices and 1280 faces and render at 64$\times$64. While PyTorch3D performs better under that setting, we see that the absolute performance for both SoftRas (chamfer 1.475) and PyTorch3D (chamfer 0.989) is not good enough. Going to higher rendering resolutions improves models (chamfer 0.346 with SoftRas vs 0.313 with PyTorch3D). Sphere GCN, a more geometry-aware network architecture, performs much better for all rendering resolutions and all rendering engines (chamfer 0.301 with SoftRas vs chamfer 0.293 with PyTorch3D at 128$\times$128). This is also evident in Figure~\ref{fig:silh_mesh_render} where we show predictions for all models at a 128$\times$128 rendering resolution. Sphere FC is able to capture a global pose and appearance but fails to capture instance-specific shape details. Sphere GCN, which deforms the same sized sphere template, is able to reconstruct instance-specific shapes, such as chair legs and tables, much more accurately. We take advantage of PyTorch3D's scale efficiency and use a larger sphere where we immediately see improved reconstruction quality (chamfer 0.281 with PyTorch3D). However, Sphere FC and Sphere GCN can only make predictions homeomorphic to spheres. This forces them to multiple face intersections and irregularly shaped faces in order to capture the complex shape topologies. Our Voxel GCN variant tests the ability of our renderer to make predictions of any genus by refining shape topologies predicted by a neural network. The reconstruction quality improves for Voxel GCN which captures holes and complex shapes while maintaining regular shaped faces. More importantly, Voxel GCN improves the Voxel Only baseline, which also predicts voxels but performs no further refinement. Even though both models are trained with the same supervision, Voxel GCN achieves a chamfer of 0.267 compared to 0.916 for Voxel Only.

\begin{figure}
  \centering
  \begin{minipage}{0.1\linewidth}\centering Input \end{minipage}
  \begin{minipage}{0.2\linewidth}\centering Sphere FC \end{minipage}
  \begin{minipage}{0.2\linewidth}\centering Sphere GCN \end{minipage}
   \begin{minipage}{0.2\linewidth}\centering High Res \\ Sphere GCN \end{minipage}
  \begin{minipage}{0.2\linewidth}\centering Voxel GCN \end{minipage} \\*
  \includegraphics[width=0.1\linewidth]{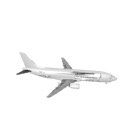}
  \includegraphics[width=0.1\linewidth]{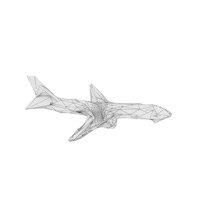}
  \includegraphics[width=0.1\linewidth]{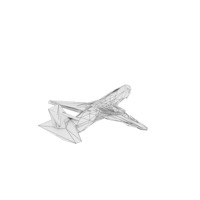}
  \includegraphics[width=0.1\linewidth]{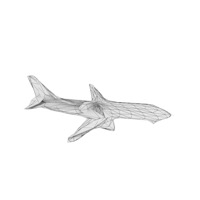}
  \includegraphics[width=0.1\linewidth]{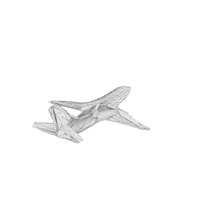}
  \includegraphics[width=0.1\linewidth]{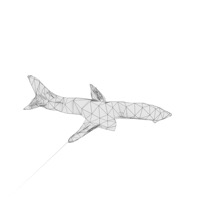}
  \includegraphics[width=0.1\linewidth]{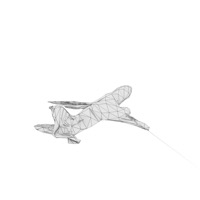}
  \includegraphics[width=0.1\linewidth]{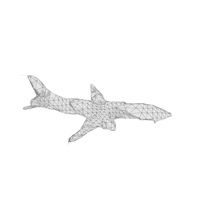}
  \includegraphics[width=0.1\linewidth]{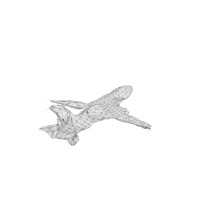} \\
  
  \includegraphics[width=0.1\linewidth]{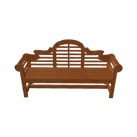}
  \includegraphics[width=0.1\linewidth]{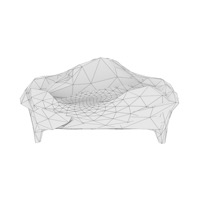}
  \includegraphics[width=0.1\linewidth]{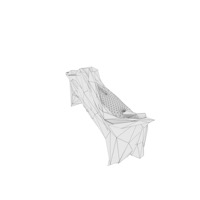}
  \includegraphics[width=0.1\linewidth]{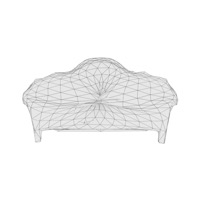}
  \includegraphics[width=0.1\linewidth]{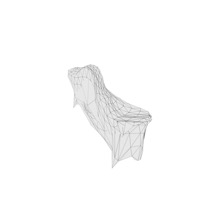}
  \includegraphics[width=0.1\linewidth]{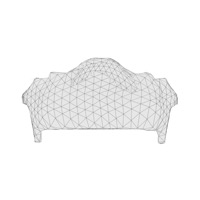}
  \includegraphics[width=0.1\linewidth]{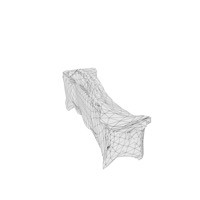}
  \includegraphics[width=0.1\linewidth]{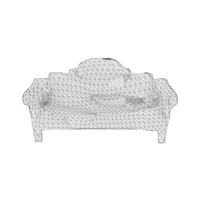}
  \includegraphics[width=0.1\linewidth]{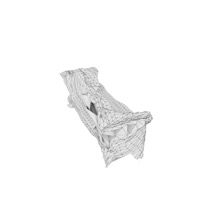} \\
  
  \includegraphics[width=0.1\linewidth]{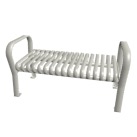}
  \includegraphics[width=0.1\linewidth]{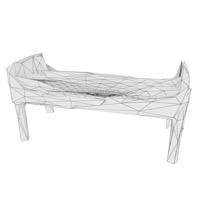}
  \includegraphics[width=0.1\linewidth]{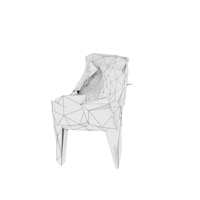}
  \includegraphics[width=0.1\linewidth]{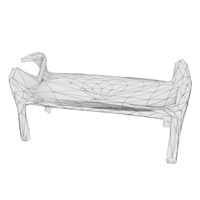}
  \includegraphics[width=0.1\linewidth]{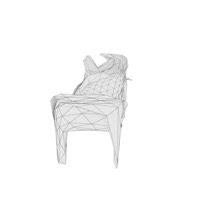}
  \includegraphics[width=0.1\linewidth]{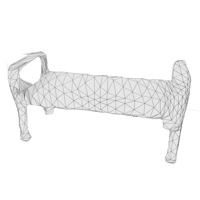}
  \includegraphics[width=0.1\linewidth]{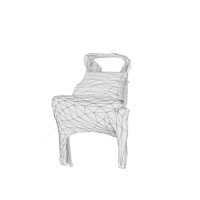}
  \includegraphics[width=0.1\linewidth]{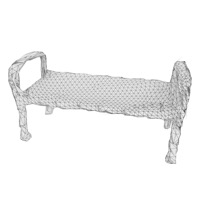}
  \includegraphics[width=0.1\linewidth]{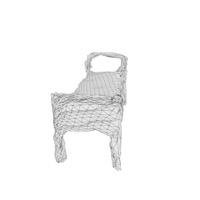} \\

  \includegraphics[width=0.1\linewidth]{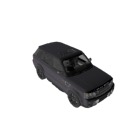}
  \includegraphics[width=0.1\linewidth]{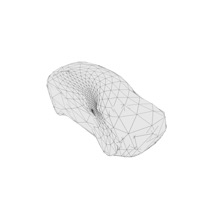}
  \includegraphics[width=0.1\linewidth]{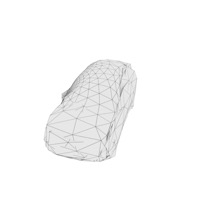}
  \includegraphics[width=0.1\linewidth]{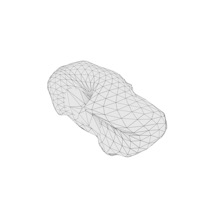}
  \includegraphics[width=0.1\linewidth]{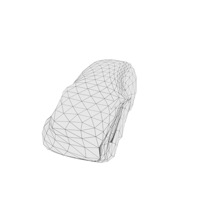}
   \includegraphics[width=0.1\linewidth]{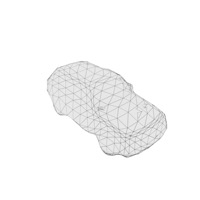}
  \includegraphics[width=0.1\linewidth]{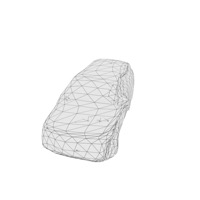}
  \includegraphics[width=0.1\linewidth]{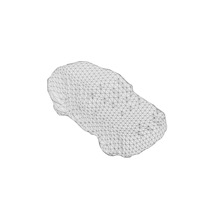}
  \includegraphics[width=0.1\linewidth]{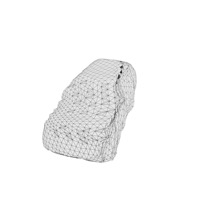} \\

  \includegraphics[width=0.1\linewidth]{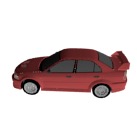}
  \includegraphics[width=0.1\linewidth]{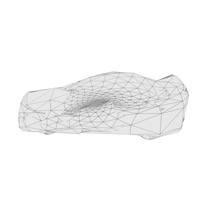}
  \includegraphics[width=0.1\linewidth]{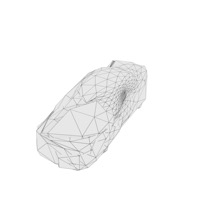}
  \includegraphics[width=0.1\linewidth]{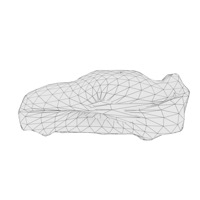}
  \includegraphics[width=0.1\linewidth]{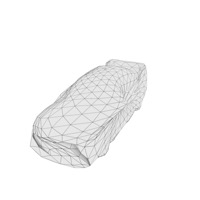}
   \includegraphics[width=0.1\linewidth]{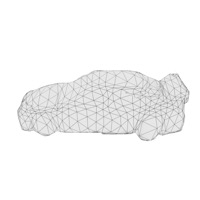}
  \includegraphics[width=0.1\linewidth]{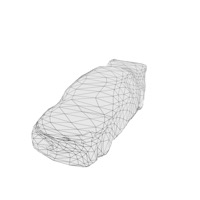}
  \includegraphics[width=0.1\linewidth]{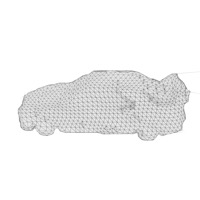}
  \includegraphics[width=0.1\linewidth]{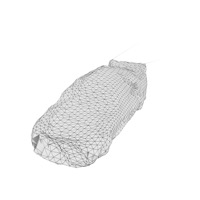} \\
  
  \includegraphics[width=0.1\linewidth]{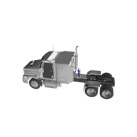}
  \includegraphics[width=0.1\linewidth]{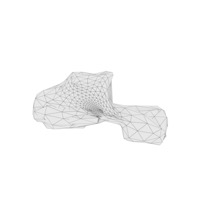}
  \includegraphics[width=0.1\linewidth]{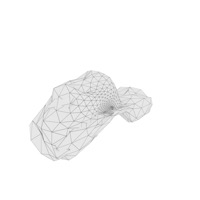}
  \includegraphics[width=0.1\linewidth]{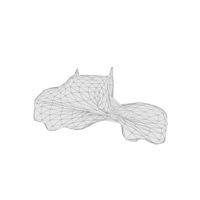}
  \includegraphics[width=0.1\linewidth]{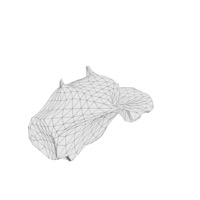}
   \includegraphics[width=0.1\linewidth]{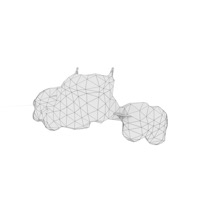}
  \includegraphics[width=0.1\linewidth]{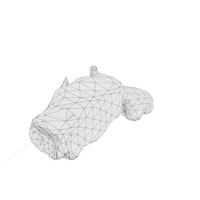}
  \includegraphics[width=0.1\linewidth]{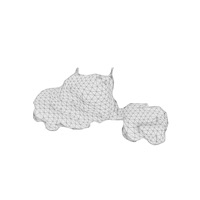}
  \includegraphics[width=0.1\linewidth]{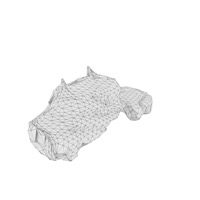} \\
  
  \includegraphics[width=0.1\linewidth]{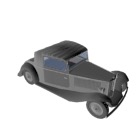}
  \includegraphics[width=0.1\linewidth]{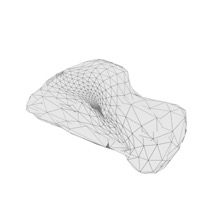}
  \includegraphics[width=0.1\linewidth]{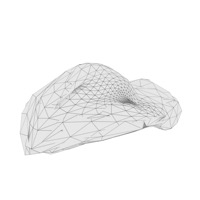}
  \includegraphics[width=0.1\linewidth]{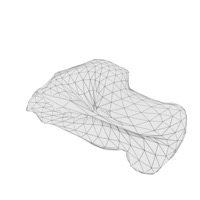}
  \includegraphics[width=0.1\linewidth]{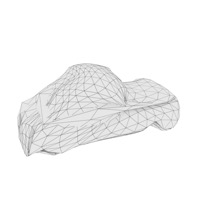}
   \includegraphics[width=0.1\linewidth]{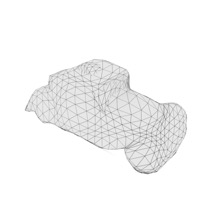}
  \includegraphics[width=0.1\linewidth]{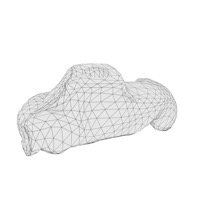}
  \includegraphics[width=0.1\linewidth]{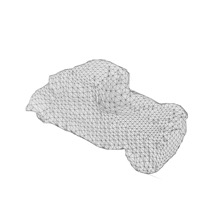}
  \includegraphics[width=0.1\linewidth]{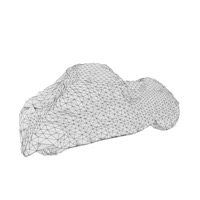} \\
  
  \includegraphics[width=0.1\linewidth]{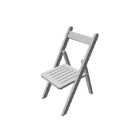}
  \includegraphics[width=0.1\linewidth]{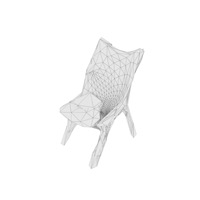}
  \includegraphics[width=0.1\linewidth]{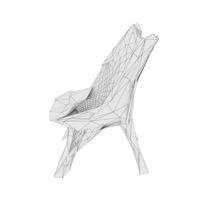}
  \includegraphics[width=0.1\linewidth]{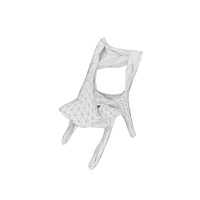}
  \includegraphics[width=0.1\linewidth]{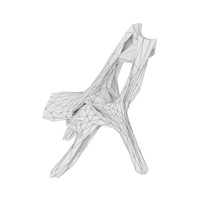}
   \includegraphics[width=0.1\linewidth]{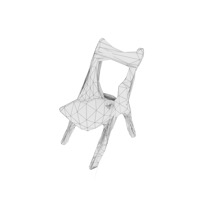}
  \includegraphics[width=0.1\linewidth]{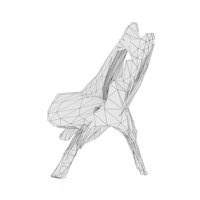}
  \includegraphics[width=0.1\linewidth]{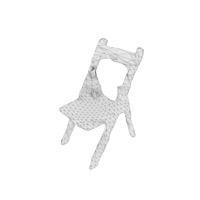}
  \includegraphics[width=0.1\linewidth]{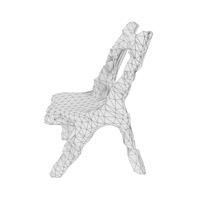} \\
  
  \includegraphics[width=0.1\linewidth]{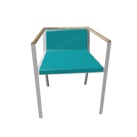}
  \includegraphics[width=0.1\linewidth]{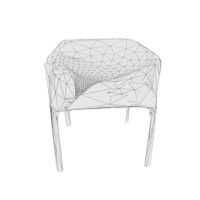}
  \includegraphics[width=0.1\linewidth]{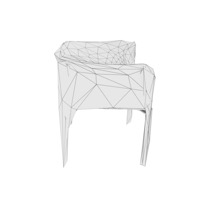}
  \includegraphics[width=0.1\linewidth]{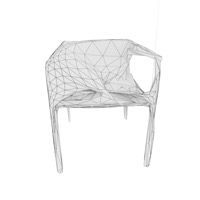}
  \includegraphics[width=0.1\linewidth]{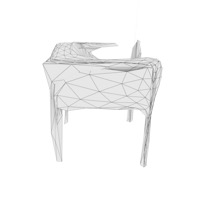}
   \includegraphics[width=0.1\linewidth]{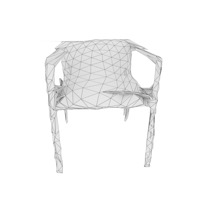}
  \includegraphics[width=0.1\linewidth]{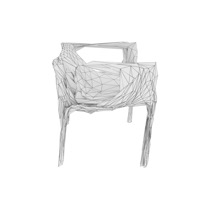}
  \includegraphics[width=0.1\linewidth]{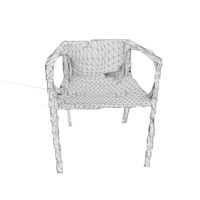}
  \includegraphics[width=0.1\linewidth]{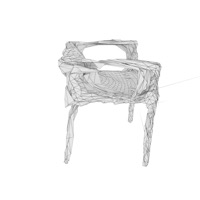} \\
  
  \includegraphics[width=0.1\linewidth]{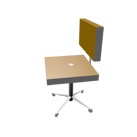}
  \includegraphics[width=0.1\linewidth]{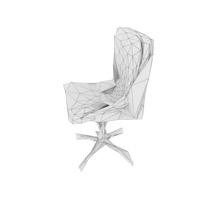}
  \includegraphics[width=0.1\linewidth]{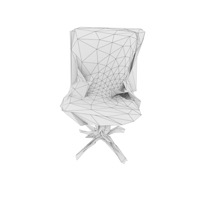}
  \includegraphics[width=0.1\linewidth]{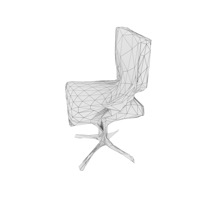}
  \includegraphics[width=0.1\linewidth]{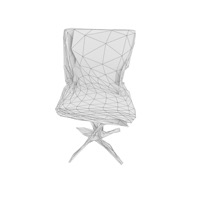}
   \includegraphics[width=0.1\linewidth]{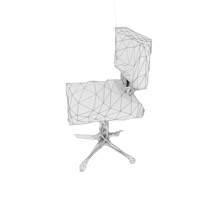}
  \includegraphics[width=0.1\linewidth]{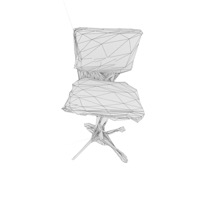}
  \includegraphics[width=0.1\linewidth]{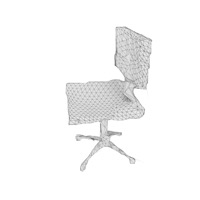}
  \includegraphics[width=0.1\linewidth]{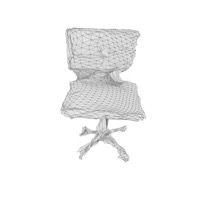} \\
  
  \includegraphics[width=0.1\linewidth]{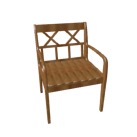}
  \includegraphics[width=0.1\linewidth]{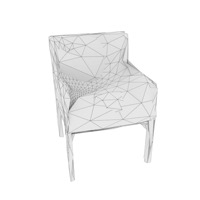}
  \includegraphics[width=0.1\linewidth]{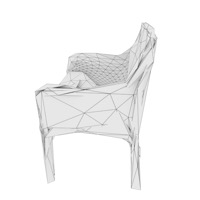}
  \includegraphics[width=0.1\linewidth]{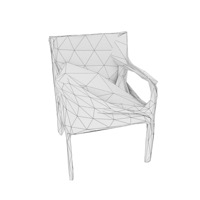}
  \includegraphics[width=0.1\linewidth]{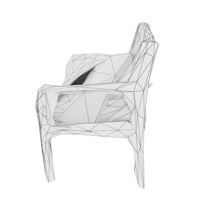}
   \includegraphics[width=0.1\linewidth]{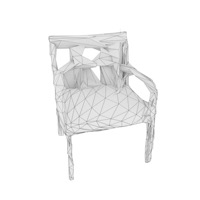}
  \includegraphics[width=0.1\linewidth]{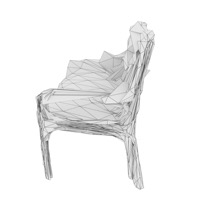}
  \includegraphics[width=0.1\linewidth]{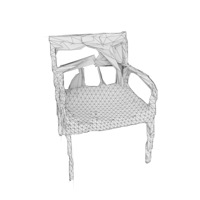}
  \includegraphics[width=0.1\linewidth]{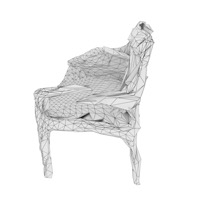} \\
  
  \includegraphics[width=0.1\linewidth]{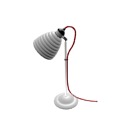}
  \includegraphics[width=0.1\linewidth]{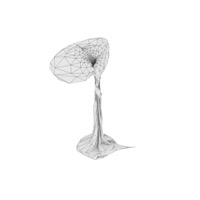}
  \includegraphics[width=0.1\linewidth]{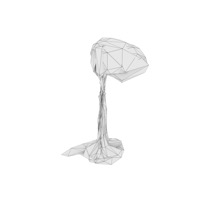}
  \includegraphics[width=0.1\linewidth]{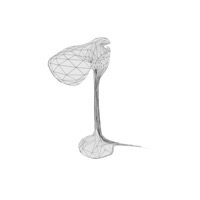}
  \includegraphics[width=0.1\linewidth]{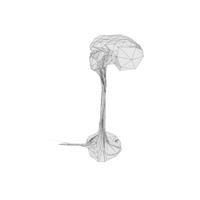}
   \includegraphics[width=0.1\linewidth]{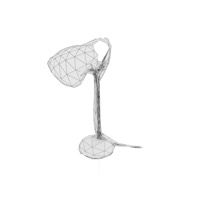}
  \includegraphics[width=0.1\linewidth]{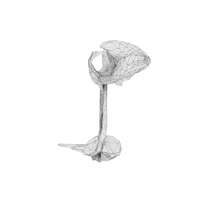}
  \includegraphics[width=0.1\linewidth]{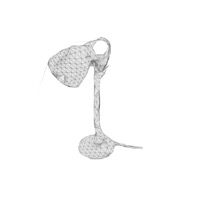}
  \includegraphics[width=0.1\linewidth]{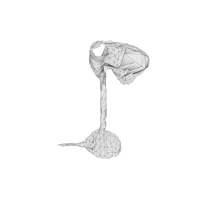} \\
  
  \includegraphics[width=0.1\linewidth]{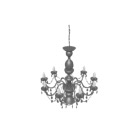}
  \includegraphics[width=0.1\linewidth]{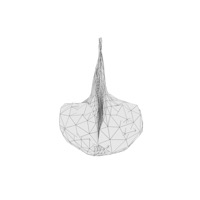}
  \includegraphics[width=0.1\linewidth]{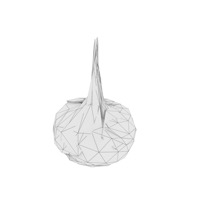}
  \includegraphics[width=0.1\linewidth]{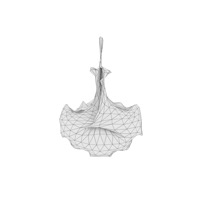}
  \includegraphics[width=0.1\linewidth]{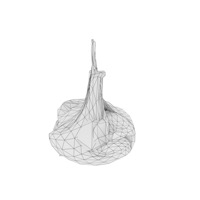}
   \includegraphics[width=0.1\linewidth]{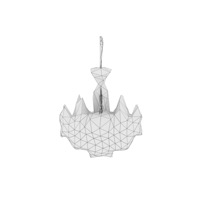}
  \includegraphics[width=0.1\linewidth]{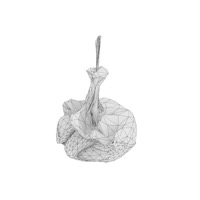}
  \includegraphics[width=0.1\linewidth]{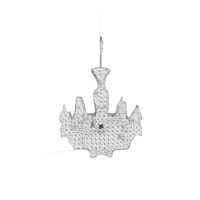}
  \includegraphics[width=0.1\linewidth]{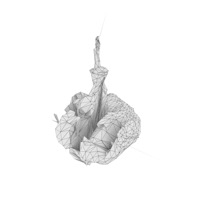} \\
  
  \includegraphics[width=0.1\linewidth]{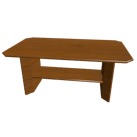}
  \includegraphics[width=0.1\linewidth]{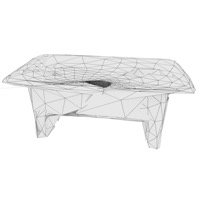}
  \includegraphics[width=0.1\linewidth]{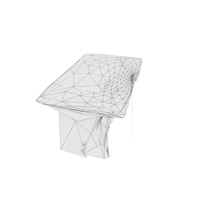}
  \includegraphics[width=0.1\linewidth]{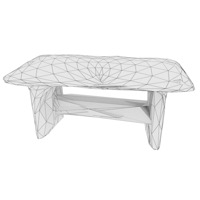}
  \includegraphics[width=0.1\linewidth]{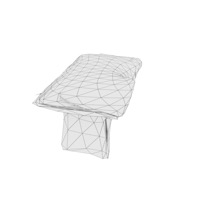}
   \includegraphics[width=0.1\linewidth]{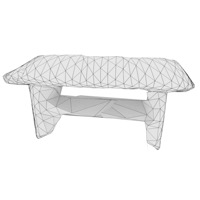}
  \includegraphics[width=0.1\linewidth]{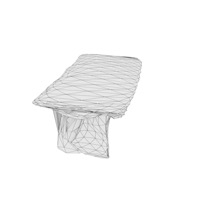}
  \includegraphics[width=0.1\linewidth]{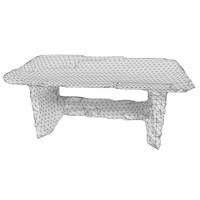}
  \includegraphics[width=0.1\linewidth]{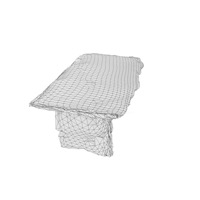} \\
  
   \includegraphics[width=0.1\linewidth]{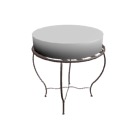}
  \includegraphics[width=0.1\linewidth]{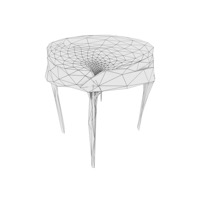}
  \includegraphics[width=0.1\linewidth]{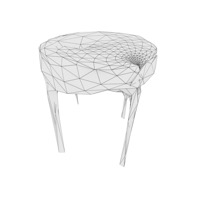}
  \includegraphics[width=0.1\linewidth]{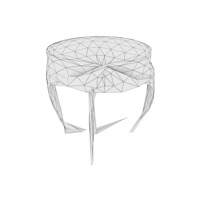}
  \includegraphics[width=0.1\linewidth]{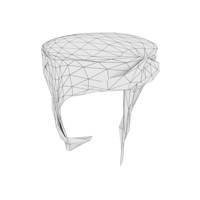}
   \includegraphics[width=0.1\linewidth]{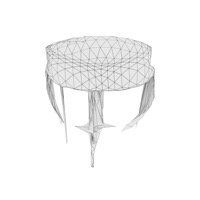}
  \includegraphics[width=0.1\linewidth]{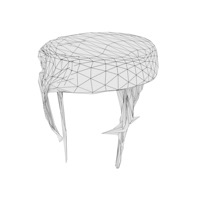}
  \includegraphics[width=0.1\linewidth]{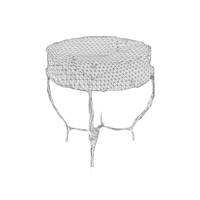}
  \includegraphics[width=0.1\linewidth]{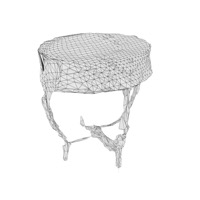} \\

  \caption{Shape reconstruction predictions via silhouette mesh rendering on ShapeNet test. We show the input image (left). For each model, we show the object shape prediction made the model and an additional random view.}
  \label{fig:silh_mesh_render_appendix}
\end{figure}

\begin{figure}
  \begin{subfigure}{0.48\linewidth}
  \centering
  \begin{minipage}{0.18\linewidth} \centering Input \end{minipage}
  \begin{minipage}{0.36\linewidth}\centering Sphere GCN \end{minipage} 
  \begin{minipage}{0.36\linewidth}\centering  Voxel GCN \end{minipage}
  
    \includegraphics[width=0.18\linewidth]{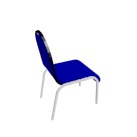}
  \begin{minipage}[t][0.18\linewidth][t]{0.36\linewidth}
  	\centering
  	\includegraphics[width=0.5\linewidth]{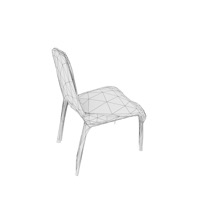}
  \end{minipage}
   \begin{minipage}[t][0.18\linewidth][t]{0.36\linewidth}
  	\centering
  	\includegraphics[width=0.5\linewidth]{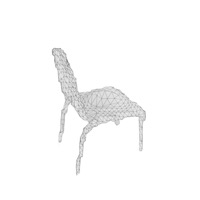} 
  \end{minipage} \\*
  
 \FramedBox{0.18\linewidth}{0.18\linewidth}{\small Flat}
 \includegraphics[width=0.18\linewidth]{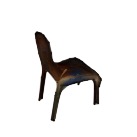}
 \includegraphics[width=0.18\linewidth]{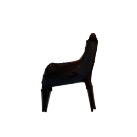}
 \includegraphics[width=0.18\linewidth]{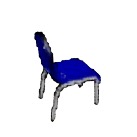}
 \includegraphics[width=0.18\linewidth]{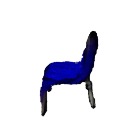} 
  
 \FramedBox{0.18\linewidth}{0.18\linewidth}{\small Phong}
 \includegraphics[width=0.18\linewidth]{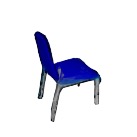}
 \includegraphics[width=0.18\linewidth]{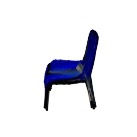}
 \includegraphics[width=0.18\linewidth]{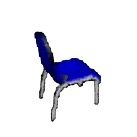}
 \includegraphics[width=0.18\linewidth]{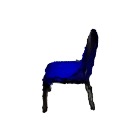} 
 
 \FramedBox{0.18\linewidth}{0.18\linewidth}{\small Gouraud}
 \includegraphics[width=0.18\linewidth]{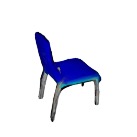}
 \includegraphics[width=0.18\linewidth]{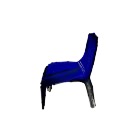}
 \includegraphics[width=0.18\linewidth]{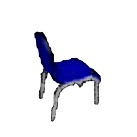}
 \includegraphics[width=0.18\linewidth]{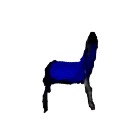} 
 \end{subfigure}
  \begin{subfigure}{0.48\linewidth}
  \centering
  \begin{minipage}{0.18\linewidth} \centering Input \end{minipage}
  \begin{minipage}{0.36\linewidth}\centering Sphere GCN \end{minipage} 
  \begin{minipage}{0.36\linewidth}\centering  Voxel GCN \end{minipage}
  
  \includegraphics[width=0.18\linewidth]{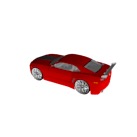}
   \begin{minipage}[t][0.18\linewidth][t]{0.36\linewidth}
  	\centering
 	 \includegraphics[width=0.5\linewidth]{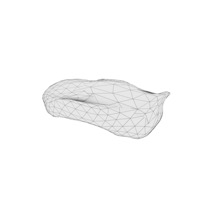}
   \end{minipage}
   \begin{minipage}[t][0.18\linewidth][t]{0.36\linewidth}
   	\centering
    	\includegraphics[width=0.5\linewidth]{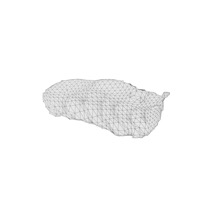}
   \end{minipage}\\
  
  \FramedBox{0.18\linewidth}{0.18\linewidth}{\small Flat}
  \includegraphics[width=0.18\linewidth]{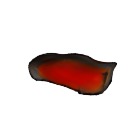}
  \includegraphics[width=0.18\linewidth]{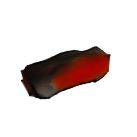}
  \includegraphics[width=0.18\linewidth]{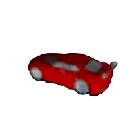}
  \includegraphics[width=0.18\linewidth]{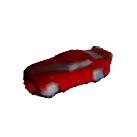} 
 
  \FramedBox{0.18\linewidth}{0.18\linewidth}{\small Phong}
  \includegraphics[width=0.18\linewidth]{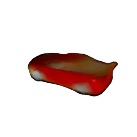}
  \includegraphics[width=0.18\linewidth]{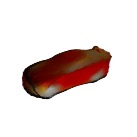}
  \includegraphics[width=0.18\linewidth]{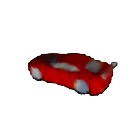}
  \includegraphics[width=0.18\linewidth]{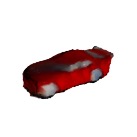} 
 
  \FramedBox{0.18\linewidth}{0.18\linewidth}{\small Gouraud}
  \includegraphics[width=0.18\linewidth]{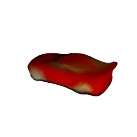}
  \includegraphics[width=0.18\linewidth]{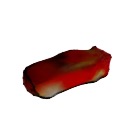}
  \includegraphics[width=0.18\linewidth]{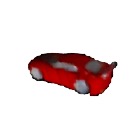}
  \includegraphics[width=0.18\linewidth]{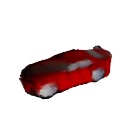} 
  \end{subfigure} 
  \begin{subfigure}{0.48\linewidth}
  \centering
  \includegraphics[width=0.18\linewidth]{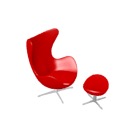}
   \begin{minipage}[t][0.18\linewidth][t]{0.36\linewidth}
  	\centering
 	 \includegraphics[width=0.5\linewidth]{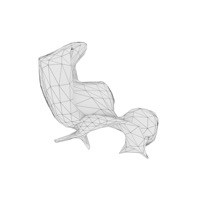}
   \end{minipage}
   \begin{minipage}[t][0.18\linewidth][t]{0.36\linewidth}
   	\centering
    	\includegraphics[width=0.5\linewidth]{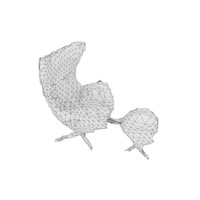}
   \end{minipage}\\
  
  \FramedBox{0.18\linewidth}{0.18\linewidth}{\small Flat}
  \includegraphics[width=0.18\linewidth]{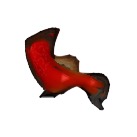}
  \includegraphics[width=0.18\linewidth]{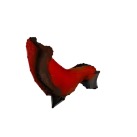}
  \includegraphics[width=0.18\linewidth]{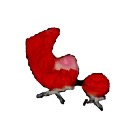}
  \includegraphics[width=0.18\linewidth]{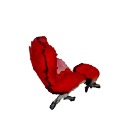} 
 
  \FramedBox{0.18\linewidth}{0.18\linewidth}{\small Phong}
  \includegraphics[width=0.18\linewidth]{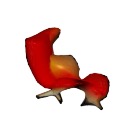}
  \includegraphics[width=0.18\linewidth]{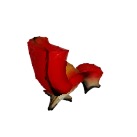}
  \includegraphics[width=0.18\linewidth]{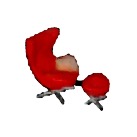}
  \includegraphics[width=0.18\linewidth]{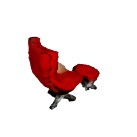} 
 
  \FramedBox{0.18\linewidth}{0.18\linewidth}{\small Gouraud}
  \includegraphics[width=0.18\linewidth]{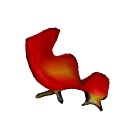}
  \includegraphics[width=0.18\linewidth]{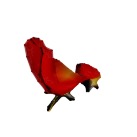}
  \includegraphics[width=0.18\linewidth]{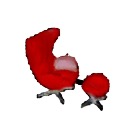}
  \includegraphics[width=0.18\linewidth]{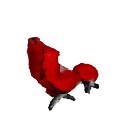} 
  \end{subfigure}
  \hspace{5mm}
  \begin{subfigure}{0.48\linewidth}
  \centering
  \includegraphics[width=0.18\linewidth]{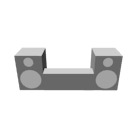}
   \begin{minipage}[t][0.18\linewidth][t]{0.36\linewidth}
  	\centering
 	 \includegraphics[width=0.5\linewidth]{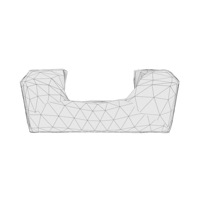}
   \end{minipage}
   \begin{minipage}[t][0.18\linewidth][t]{0.36\linewidth}
   	\centering
    	\includegraphics[width=0.5\linewidth]{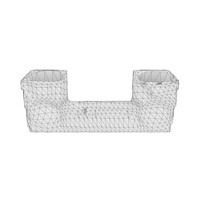}
   \end{minipage}\\
  
  \FramedBox{0.18\linewidth}{0.18\linewidth}{\small Flat}
  \includegraphics[width=0.18\linewidth]{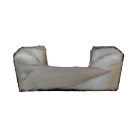}
  \includegraphics[width=0.18\linewidth]{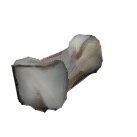}
  \includegraphics[width=0.18\linewidth]{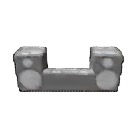}
  \includegraphics[width=0.18\linewidth]{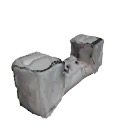} 
 
  \FramedBox{0.18\linewidth}{0.18\linewidth}{\small Phong}
  \includegraphics[width=0.18\linewidth]{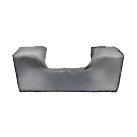}
  \includegraphics[width=0.18\linewidth]{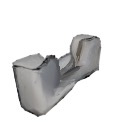}
  \includegraphics[width=0.18\linewidth]{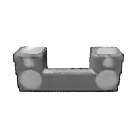}
  \includegraphics[width=0.18\linewidth]{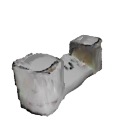} 
 
  \FramedBox{0.18\linewidth}{0.18\linewidth}{\small Gouraud}
  \includegraphics[width=0.18\linewidth]{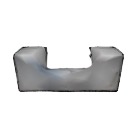}
  \includegraphics[width=0.18\linewidth]{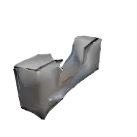}
  \includegraphics[width=0.18\linewidth]{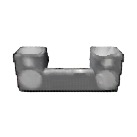}
  \includegraphics[width=0.18\linewidth]{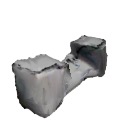} 
  \end{subfigure}
  \caption{Shape and texture predictions via textured mesh rendering on the ShapeNet test set. We show the input image (left). For each shading function, we show the shape prediction and its textured rendering (with an additional view) as predicted by the model.}
  \label{fig:text_mesh_render_appendix}
\end{figure}

\mypar{More visualizations}
We show additional mesh predictions via silhouette rendering in Figure~\ref{fig:silh_mesh_render_appendix} and via textured rendering in Figure~\ref{fig:text_mesh_render_appendix}. Our texture model is simple; we predict $(r,g,b)$ texture values per vertex. The shaders interpolate the vertex textures into face textures based on their algorithm and the blending accumulates the face textures for each pixel. More sophisticated texture models could involve directly sampling vertex textures from the input image or using GANs for natural looking texture predictions. Despite its simplicity, our texture model does well at reconstructing textures. Textures are more accurate for Voxel GCN mainly because of the regular shaped and sized faces in the predicted mesh. This is in contrast to Sphere GCN, where faces can intersect and largely vary in size in an effort to capture the complex object shape, which in turn affects the interpolated face textures. In addition, simple shading functions, such as \emph{flat shading}, visibly lead to worse textures, as expected. More sophisticated shaders, like Phong and Gouraud, lead to better texture reconstructions. 

\subsection{Unsupervised point cloud prediction}
\begin{figure}
 \centering
 \includegraphics[width=0.99\linewidth]{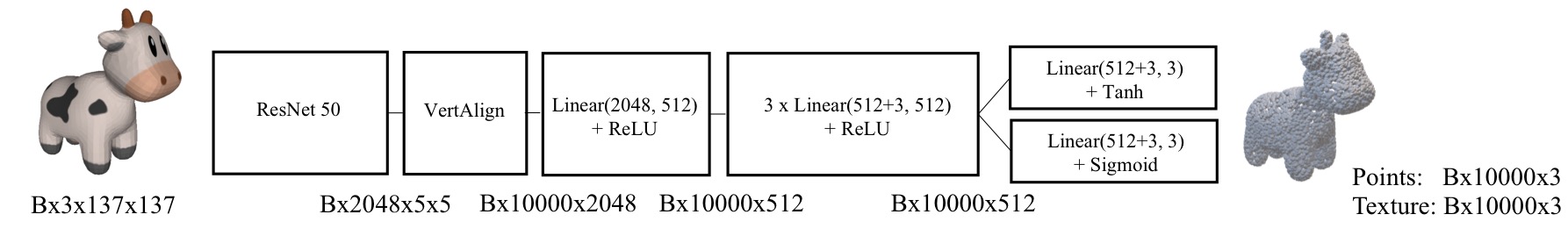}
  \caption{The Point Align network architecture used for unsupervised point cloud predictions in our experiments.}
  \label{fig:point_networks}
\end{figure}

\begin{figure}
  \begin{subfigure}{0.5\linewidth}
  \centering
  \begin{minipage}{0.18\linewidth}\centering Input \end{minipage}
  \begin{minipage}{0.36\linewidth}\centering Alpha \end{minipage}
  \begin{minipage}{0.36\linewidth}\centering Norm \end{minipage}
   \includegraphics[width=0.18\linewidth]{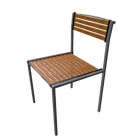}
  \includegraphics[width=0.18\linewidth]{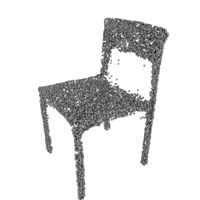}   
  \includegraphics[width=0.18\linewidth]{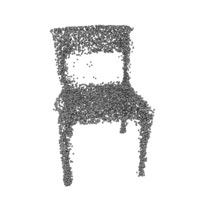} 
  \includegraphics[width=0.18\linewidth]{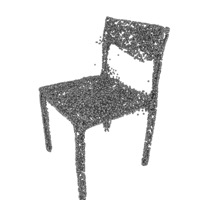}
  \includegraphics[width=0.18\linewidth]{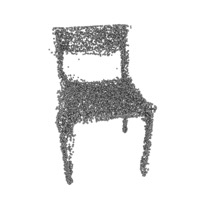} \\
 
  \FramedBox{0.18\linewidth}{0.18\linewidth}{\textcolor{white}{Point Align}}
  \includegraphics[width=0.18\linewidth]{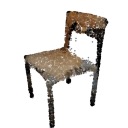}   
  \includegraphics[width=0.18\linewidth]{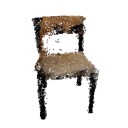} 
  \includegraphics[width=0.18\linewidth]{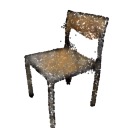}
  \includegraphics[width=0.18\linewidth]{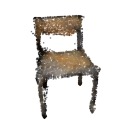} \\
  \end{subfigure}
  \begin{subfigure}{0.5\linewidth}
  \centering
  \begin{minipage}{0.18\linewidth}\centering Input \end{minipage}
  \begin{minipage}{0.36\linewidth}\centering Alpha \end{minipage}
  \begin{minipage}{0.36\linewidth}\centering Norm \end{minipage}
   \includegraphics[width=0.18\linewidth]{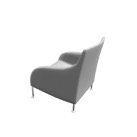}
  \includegraphics[width=0.18\linewidth]{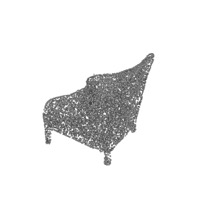}   
  \includegraphics[width=0.18\linewidth]{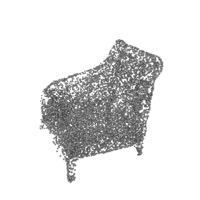} 
  \includegraphics[width=0.18\linewidth]{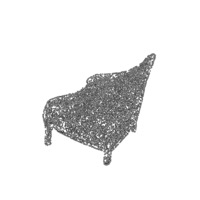}
  \includegraphics[width=0.18\linewidth]{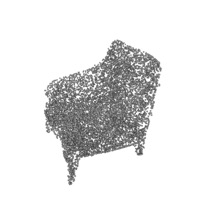} \\
 
  \FramedBox{0.18\linewidth}{0.18\linewidth}{\textcolor{white}{Point Align}}
  \includegraphics[width=0.18\linewidth]{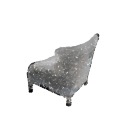}   
  \includegraphics[width=0.18\linewidth]{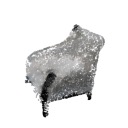} 
  \includegraphics[width=0.18\linewidth]{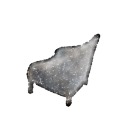}
  \includegraphics[width=0.18\linewidth]{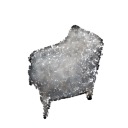} \\
  \end{subfigure}
  \begin{subfigure}{0.5\linewidth}
  \centering
  \includegraphics[width=0.18\linewidth]{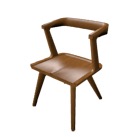}
  \includegraphics[width=0.18\linewidth]{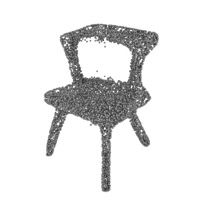}   
  \includegraphics[width=0.18\linewidth]{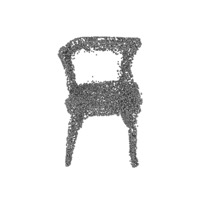} 
  \includegraphics[width=0.18\linewidth]{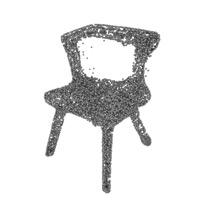}
  \includegraphics[width=0.18\linewidth]{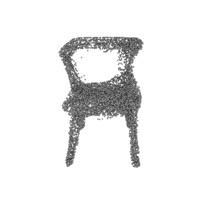} \\
 
  \FramedBox{0.18\linewidth}{0.18\linewidth}{\textcolor{white}{Point Align}}
  \includegraphics[width=0.18\linewidth]{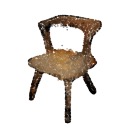}   
  \includegraphics[width=0.18\linewidth]{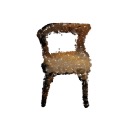} 
  \includegraphics[width=0.18\linewidth]{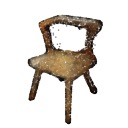}
  \includegraphics[width=0.18\linewidth]{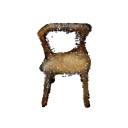} \\
  \end{subfigure}
  \begin{subfigure}{0.5\linewidth}
  \centering
   \includegraphics[width=0.18\linewidth]{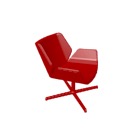}
  \includegraphics[width=0.18\linewidth]{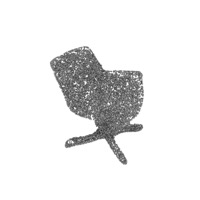}   
  \includegraphics[width=0.18\linewidth]{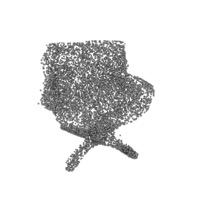} 
  \includegraphics[width=0.18\linewidth]{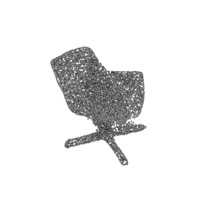}
  \includegraphics[width=0.18\linewidth]{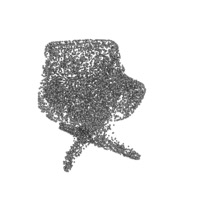} \\
 
  \FramedBox{0.18\linewidth}{0.18\linewidth}{\textcolor{white}{Point Align}}
  \includegraphics[width=0.18\linewidth]{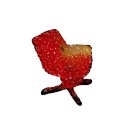}   
  \includegraphics[width=0.18\linewidth]{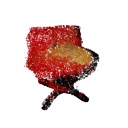} 
  \includegraphics[width=0.18\linewidth]{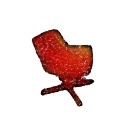}
  \includegraphics[width=0.18\linewidth]{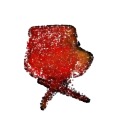} \\
  \end{subfigure}
  \begin{subfigure}{0.5\linewidth}
  \centering
   \includegraphics[width=0.18\linewidth]{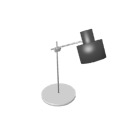}
  \includegraphics[width=0.18\linewidth]{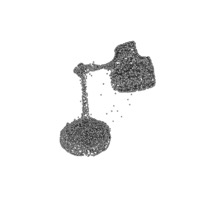}   
  \includegraphics[width=0.18\linewidth]{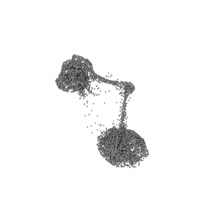} 
  \includegraphics[width=0.18\linewidth]{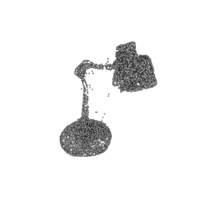}
  \includegraphics[width=0.18\linewidth]{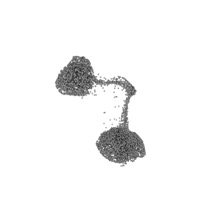} \\
 
  \FramedBox{0.18\linewidth}{0.18\linewidth}{\textcolor{white}{Point Align}}
  \includegraphics[width=0.18\linewidth]{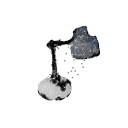}   
  \includegraphics[width=0.18\linewidth]{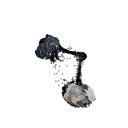} 
  \includegraphics[width=0.18\linewidth]{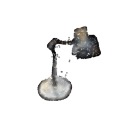}
  \includegraphics[width=0.18\linewidth]{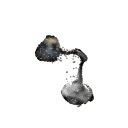} \\
  \end{subfigure}
  \begin{subfigure}{0.5\linewidth}
  \centering
   \includegraphics[width=0.18\linewidth]{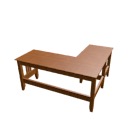}
  \includegraphics[width=0.18\linewidth]{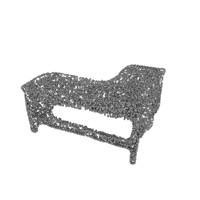}   
  \includegraphics[width=0.18\linewidth]{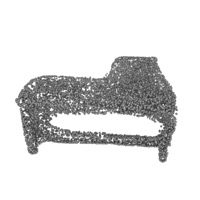} 
  \includegraphics[width=0.18\linewidth]{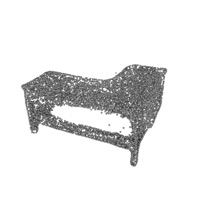}
  \includegraphics[width=0.18\linewidth]{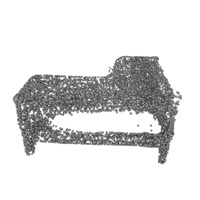} \\
 
  \FramedBox{0.18\linewidth}{0.18\linewidth}{\textcolor{white}{Point Align}}
  \includegraphics[width=0.18\linewidth]{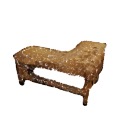}   
  \includegraphics[width=0.18\linewidth]{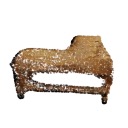} 
  \includegraphics[width=0.18\linewidth]{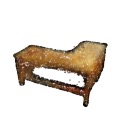}
  \includegraphics[width=0.18\linewidth]{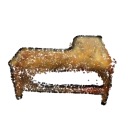} \\
  \end{subfigure}
  \begin{subfigure}{0.5\linewidth}
  \centering
   \includegraphics[width=0.18\linewidth]{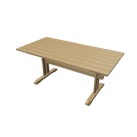}
  \includegraphics[width=0.18\linewidth]{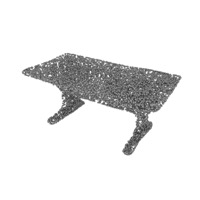}   
  \includegraphics[width=0.18\linewidth]{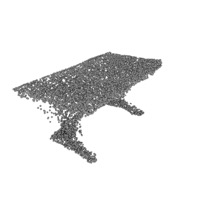} 
  \includegraphics[width=0.18\linewidth]{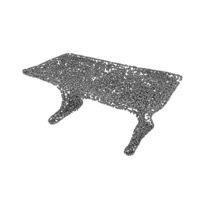}
  \includegraphics[width=0.18\linewidth]{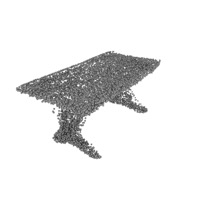} \\
 
  \FramedBox{0.18\linewidth}{0.18\linewidth}{\textcolor{white}{Point Align}}
  \includegraphics[width=0.18\linewidth]{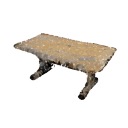}   
  \includegraphics[width=0.18\linewidth]{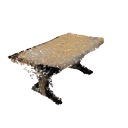} 
  \includegraphics[width=0.18\linewidth]{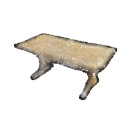}
  \includegraphics[width=0.18\linewidth]{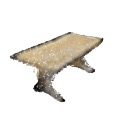} \\
  \end{subfigure}
   \begin{subfigure}{0.5\linewidth}
  \centering
   \includegraphics[width=0.18\linewidth]{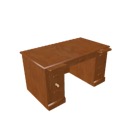}
  \includegraphics[width=0.18\linewidth]{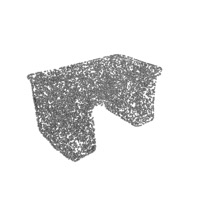}   
  \includegraphics[width=0.18\linewidth]{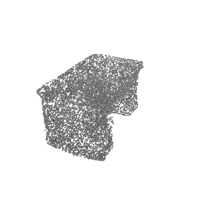} 
  \includegraphics[width=0.18\linewidth]{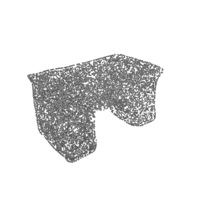}
  \includegraphics[width=0.18\linewidth]{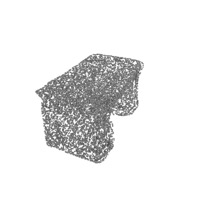} \\
 
  \FramedBox{0.18\linewidth}{0.18\linewidth}{\textcolor{white}{Point Align}}
  \includegraphics[width=0.18\linewidth]{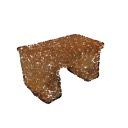}   
  \includegraphics[width=0.18\linewidth]{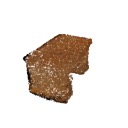} 
  \includegraphics[width=0.18\linewidth]{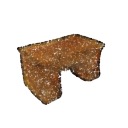}
  \includegraphics[width=0.18\linewidth]{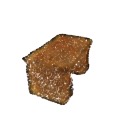} \\
  \end{subfigure}
  \caption{Point cloud and texture predictions via textured rendering on the ShapeNet test set. For each example, we show the input image and for each compositor (Alpha, Norm) we show the shape (top row) and texture (bottom row) prediction as well as an additional view.}
  \label{fig:text_cloud_render_appendix}
\end{figure}

Expanding on the brief description of the unsupervised point cloud prediction model in the main paper, here we provide more details and analyze our results further. Our training and inference procedures are similar to meshes; we follow the same 2-view train setup while at test time our model only takes as input a single RGB image (Figure~\ref{fig:overview}). Our Point Align model is similar to Sphere GCN. We start from an initial point cloud of 10k points sampled randomly and uniformly from the surface of a sphere. Each point samples features from the backbone using \valign. A sequence of fully connected layers replace \gconv  - point clouds do not have connectivity patterns - resulting in per point offset predictions (and $(r, g, b)$ values, in the case of textured models). The network architecture is shown in Figure~\ref{fig:point_networks}.

\mypar{Losses} 
Point Align is trained solely with $\mathcal{L} = \mathcal{L}_s$ for silhouette rendering and an additional $L_1$ loss between the rendered and ground truth image for textured rendering. There are no shape regularizers.

\mypar{Blending}
In the case of textured rendering, we experiment with two blending (or compositing) functions, \emph{Alpha} and \emph{Norm}. 

\mypar{More discussion on Table~\ref{tab:silh_cloud}} 
Our point cloud evaluation in Table~\ref{tab:silh_cloud} is directly comparable to that of meshes in Table~\ref{tab:silh_mesh}; we use the same number of points, 10k, to compute chamfer and $F_1$ for meshes (by sampling points from the mesh surface) and point clouds (directly using the points in the cloud). From the comparison with meshes, we observe that our unsupervised point cloud model leads to slightly better reconstruction quality than meshes (chamfer 0.272 for Point Align vs 0.281 for High Res Sphere GCN). This is proof that our point cloud renderer is effective at predicting shapes. The quality of our reconstructions is shown in Figure~\ref{fig:text_cloud_render_appendix} which provides more shape and texture predictions by our Point Align model with an alpha and norm compositor.

\bibliographystyle{abbrvnat}
\bibliography{references}

\end{document}